\documentclass[11pt,a4paper]{article}
\usepackage{iftex}
\ifpdftex
  \usepackage[utf8]{inputenc}
  \usepackage[T1]{fontenc}
  \usepackage{newtxtext,newtxmath}
  \usepackage{slantsc}
\else
  \usepackage{newtxtext,newtxmath}
\fi
\usepackage[margin=1in]{geometry}
\usepackage{graphicx}
\graphicspath{{figures/}}
\usepackage{booktabs}
\usepackage{cite}
\usepackage{array}
\usepackage{hyperref}
\usepackage[table]{xcolor} 
\usepackage{enumitem}
\usepackage{subcaption}
\usepackage{makecell}
\usepackage{tabularx}
\usepackage{arydshln}
\usepackage{float}
\usepackage{silence}
\usepackage{tablefootnote}
\usepackage{fontawesome5}
\usepackage{fancyhdr}
\usepackage{titlesec}
\usepackage{titling}
\usepackage{amsmath}
\usepackage{parskip}
\usepackage{hyphenat}
\ifpdftex
  \AtBeginDocument{%
    \fontfamily{ntxtt}\selectfont\hyphenchar\font=45\relax
    \fontfamily{txtt}\selectfont\hyphenchar\font=45\relax
    \normalfont
  }
\fi
\usepackage{listings}
\usepackage{tikz}
\usetikzlibrary{arrows.meta, positioning, shapes.geometric, spy}
\usepackage[most]{tcolorbox}
\newtcolorbox{promptbox}{
  breakable,
  colback=black!2,
  colframe=black!40,
  boxrule=0.8pt,
  arc=4pt,
  left=14pt,right=14pt,top=10pt,bottom=10pt
}
\usepackage{wrapfig}
\usepackage{algorithm}
\usepackage{algpseudocode}
\usepackage{xurl}

\usepackage{amsthm}
\newtheorem{definition}{Definition}

\definecolor{titleRed}{RGB}{170,30,45}

\hypersetup{
    colorlinks=true,
    linkcolor=blue!70!black,
    citecolor=blue!70!black,
    urlcolor=blue!70!black
}

\titleformat{\section}{\Large\bfseries}{\thesection}{0.5em}{}
\titleformat{\subsection}{\large\bfseries}{\thesubsection}{0.5em}{}
\titleformat{\subsubsection}{\normalsize\bfseries}{\thesubsubsection}{0.5em}{}
\title{\bfseries\color{titleRed}
From Skills to Talent: \\[2pt]
Organising Heterogeneous Agents as a Real-World Company}
\vspace{-3.2cm}

\author{
\textbf{Zhengxu Yu}$^{1,*}$,
\textbf{Yu Fu}$^{1,*}$,
\textbf{Zhiyuan He}$^{1}$,
\textbf{Yuxuan Huang}$^{3}$,
\textbf{Lee Ka Yiu}$^{1}$,\\
\textbf{Meng Fang}$^{3}$,
\textbf{Weilin Luo}$^{1}$,
\textbf{Jun Wang}$^{2}$
}
\begingroup
\stepcounter{footnote}%
\renewcommand{\thefootnote}{\kern0pt}%
\footnotetext[\value{footnote}]{%
\hspace{-1.8em}%
\begin{minipage}[t]{\textwidth}
\raggedright
$^{1}$HUAWEI Noah's Ark Lab \,\,
$^{2}$University College London \,\,
$^{3}$University Liverpool \,\,
$^{*}$Co-first authors \,\,
\end{minipage}%
}%
\endgroup
\renewcommand{\thefootnote}{\arabic{footnote}}
 
\date{}

\begin{document}
\vspace{-3.2cm}
\maketitle
\thispagestyle{empty}
\vspace{-1.6em}

\begin{center}
Project Homepage: \url{https://one-man-company.com} \\
Repository: \url{https://1mancompany.github.io/OneManCompany}\\
Digital Talent Market: \url{https://one-man-company.com/market}
\end{center}

\begin{abstract}
Individual agent capabilities have advanced rapidly through modular skills and 
tool integrations, yet multi-agent systems remain constrained by fixed team 
structures, tightly coupled coordination logic, and session-bound learning. We 
argue that this reflects a deeper absence: a principled organisational layer 
that governs how a workforce of agents is assembled, governed, and improved over 
time, decoupled from what individual agents know. To fill this gap, we introduce 
\emph{OneManCompany (OMC)}, a framework that elevates multi-agent systems to 
the organisational level. OMC encapsulates skills, tools, and runtime 
configurations into portable agent identities called \emph{Talents}, 
orchestrated through typed organisational interfaces that abstract over 
heterogeneous backends. A community-driven \emph{Talent Market} enables 
on-demand recruitment, allowing the organisation to close capability gaps and 
reconfigure itself dynamically during execution. Organisational decision-making 
is operationalised through an \emph{Explore-Execute-Review} ($\text{E}^2$R) 
tree search, which unifies planning, execution, and evaluation in a single 
hierarchical loop: tasks are decomposed top-down into accountable units and 
execution outcomes are aggregated bottom-up to drive systematic review and 
refinement. This loop provides formal guarantees on termination and deadlock 
freedom while mirroring the feedback mechanisms of human enterprises. Together, 
these contributions transform multi-agent systems from static, pre-configured 
pipelines into self-organising and self-improving AI organisations capable of 
adapting to open-ended tasks across diverse domains. Empirical evaluation on 
PRDBench shows that OMC achieves an $84.67\%$ success rate, surpassing the 
state of the art by $15.48$ percentage points, with cross-domain case studies 
further demonstrating its generality.
\end{abstract}

\begin{figure}[H]
\centering
\begin{tikzpicture}[
  img/.style={inner sep=0pt},
  tag/.style={font=\scriptsize\sffamily\bfseries, rounded corners=2pt, inner sep=3pt, fill=black, fill opacity=0.7, text opacity=1, text=red},
]
\node[img] (main) at (0,0) {\includegraphics[width=0.9\textwidth]{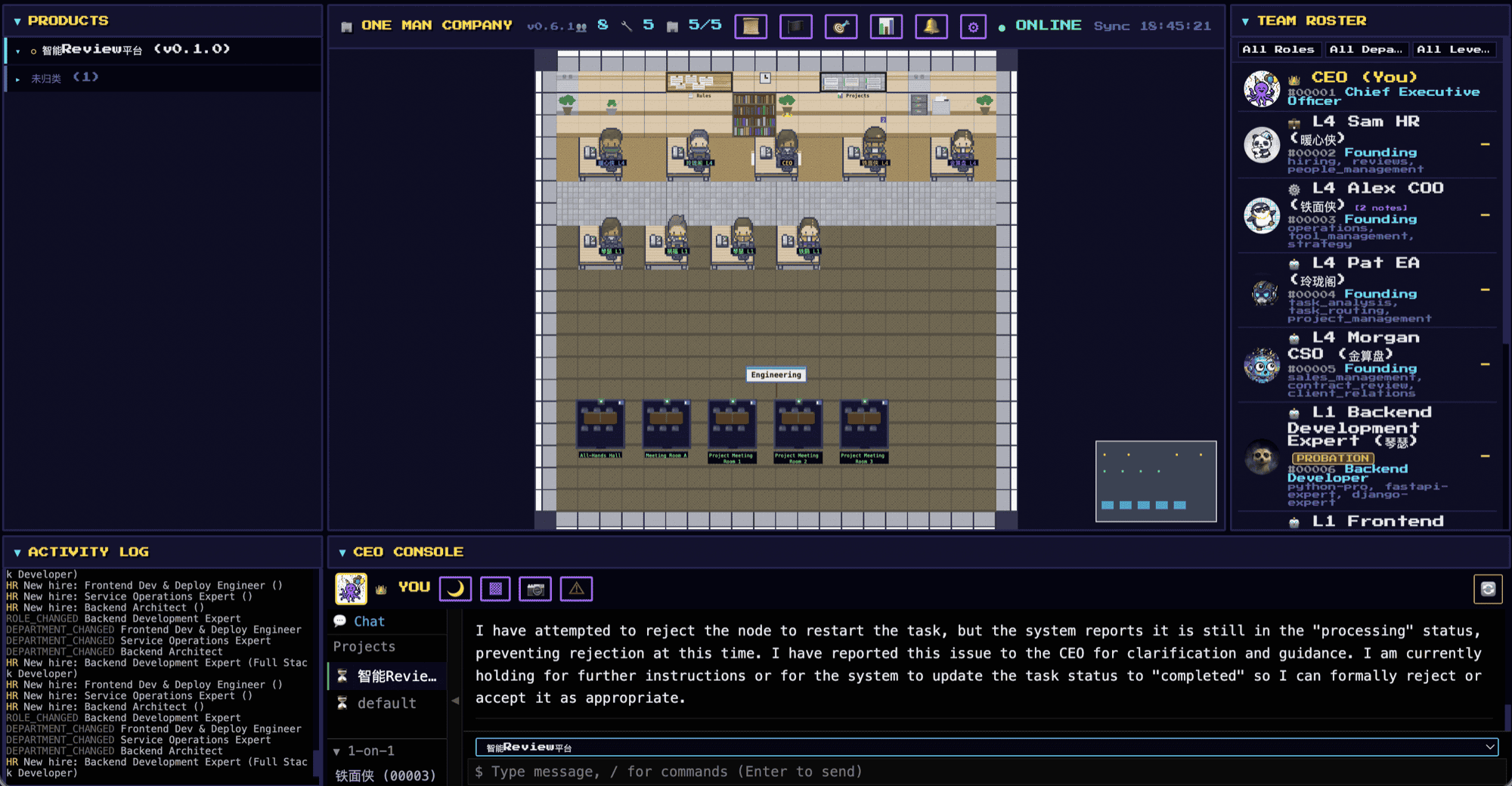}};

\node[tag] at ([xshift=5.5cm, yshift=0.5cm]main.center) {Talent Lifecycle};
\node[tag] at ([xshift=-5.8cm, yshift=2.6cm]main.center) {Task Decomposition};
\node[tag] at ([xshift=-0.2cm, yshift=-0.7cm]main.center) {Agent Coordination};
\node[tag] at ([xshift=1.5cm, yshift=3.1cm]main.center) {Org Knowledge};

\end{tikzpicture}
\caption{\textbf{The running OMC system}, where the three proposed pillars converge into a unified management interface. \emph{Talent Lifecycle} implements the Talent-Container architecture (Section~\ref{sec:harness}), with per-employee profiles tracking skills, performance, and configuration. \emph{Task Decomposition} realises the $\text{E}^2$R tree search (Section~\ref{sec:explore-review}) through hierarchical task trees with DAG dependencies. \emph{Agent Coordination} enables structured inter-agent communication (Section~\ref{sec:dag}), where agents request meetings, exchange information, and align on shared tasks through dedicated coordination channels. \emph{Org Knowledge} embodies the organisation-level evolution mechanism (Section~\ref{sec:self-evolution}), with editable workflow SOPs and company culture rules that persist across projects.}
\label{fig:overview}
\end{figure}
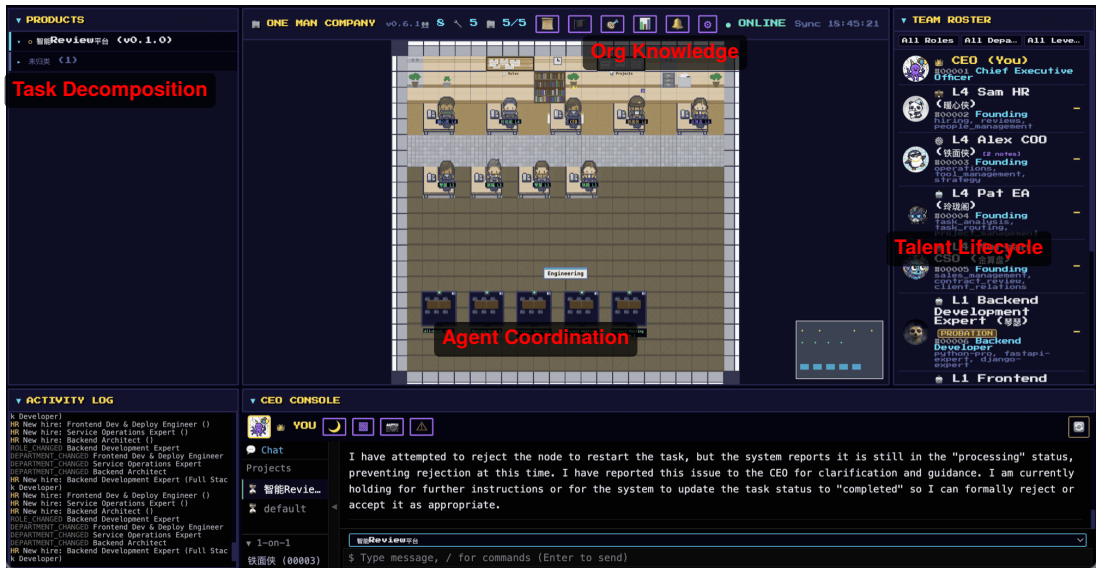

\newpage
\section{Introduction}
\label{sec:introduction}

Recent advances in large language models (LLMs) have given rise to highly capable \emph{individual AI agents}, such as Claude Code~\cite{anthropic2024claudecode}, Codex~\cite{openai2024codex}, and OpenClaw~\cite{openclaw2024}. These agents can already perform complex tasks including code generation, tool use, web interaction, and long-horizon reasoning. Much of this capability stems from a modular ecosystem of \emph{skills} and tool integrations~\cite{skillsmp, mcpzoo2025}, which allow agents to extend their functionality without modifying the underlying model. This modularity has been a key driver of progress: skills can be reused, composed, and shared across agents, enabling rapid capability accumulation at the \emph{individual} level.

However, skills operate strictly \emph{within} a single agent. They enhance what an agent \emph{can do}, but they do not address how multiple agents should \emph{work together}. As tasks grow in complexity, requiring diverse expertise, long-horizon coordination, and iterative refinement, a single agent, no matter how well equipped with skills, becomes insufficient. This has led to the emergence of \emph{multi-agent LLM systems}, where multiple agents collaborate through message passing, role assignment, or shared memory.


Despite promising progress, existing multi-agent systems remain fundamentally limited. Current orchestrator methods, e.g., CrewAI~\cite{crewai}, AutoGen~\cite{autogen}, and Paperclip~\cite{paperclip2025} either hardcode team structures which is brittle to novel projects, or let agents negotiate freely which has no convergence guarantees.  Agents from different families cannot interoperate because they are locked into incompatible runtimes.  Roles are specified through descriptive prompts rather than executable contracts, leading to hallucinated capabilities.  And self-improvement, where it exists, is session-bound and framework-specific. Recent work on \emph{dynamic agentic workflows}~\cite{wang2025tdag, erdogan2025planact, hu2025evomac} has shown that adapting task decomposition at runtime outperforms static pipelines.  But even these systems operate within a pre-configured sandbox: the team is fixed before the project starts, all agents share the same runtime, and the workflow topology is constrained to a known template. 
Essentially, these systems lack a unifying abstraction that separates \emph{organisation} from \emph{capability}. As a result, they struggle to generalise beyond narrow settings and fail to scale to open-ended, real-world projects.

We argue that the next stage of progress requires a shift in perspective: from agents and their skills to the \emph{organisation} that governs them. We define an \emph{AI organisation} as:

\vspace{10pt}
\begin{definition} [AI Organisation]
\textit{a self-governing system of heterogeneous agents with structured coordination, managed lifecycles, and experience-driven evolution.}
\end{definition}

This definition emphasises three properties absent from current paradigms. First, \emph{structured coordination}: interactions between agents are governed by explicit protocols and organisational constraints, rather than ad hoc prompting. Second, \emph{lifecycle management}: agents are created, assigned, evaluated, and retired through well-defined processes. Third, \emph{experience-driven evolution}: both agents and the organisation improve over time through systematic feedback and reflection.

This perspective fundamentally differs from both skills and conventional multi-agent systems. Skills are \emph{capability-level abstractions}, reusable within an agent but agnostic to coordination. Multi-agent systems are \emph{interaction-level abstractions}, specifying how agents communicate but not how they are organised, managed, or improved. In contrast, an AI organisation is an \emph{organisation-level abstraction}: it governs how agents are assembled into teams, how work is decomposed and executed, and how the system evolves over time. In other words, skills answer ``what can an agent do?’’, multi-agent systems answer ``how do agents interact?’’, while AI organisations answer ``how should a workforce of agents be structured and managed to achieve complex goals?’’

The absence of this organisational layer exposes a fundamental research gap:
\begin{center}
\textit{How can AI agent workforces be automatically organised, coordinated, and evolved to solve open-ended tasks across domains?}
\end{center}

We argue that this problem is analogous to organisation design in human systems, where companies provide a domain-agnostic structure for coordinating individuals, allocating resources, and improving performance over time. Importantly, this structure is decoupled from the specific knowledge of individual employees, enabling the same organisational principles to generalise across industries. We posit that a similar decoupling is essential for AI systems.

\begin{figure}[t]
    \centering
    \includegraphics[width=\textwidth]{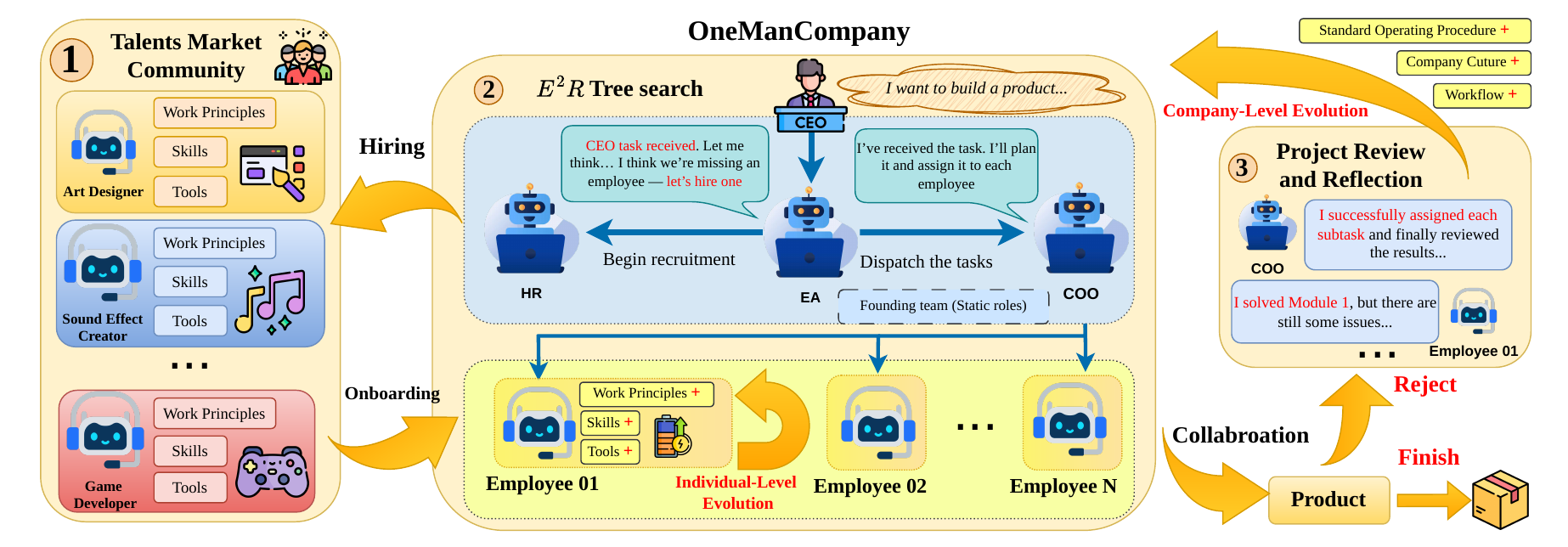}
    \caption{
An overview of the proposed OMC AI organisation system.  The central hierarchy mirrors a real company: tasks flow from the CEO or external clients to downstream AI talents, coordinated through the organisational layer.  \textit{1:} The Talent Market (left) supplies verified agents on demand.  \textit{2:} The $\text{E}^2$R tree search (centre) decomposes projects into structured task trees with dependency tracking.  \textit{3:} Individual-Level Evolution (bottom) enables agents to self-improve through tasks, while Project review (right) supports collective reflection and drives Organisation-Level Evolution.
}
    \label{fig:system-ui}
\end{figure}

To address this gap, we introduce OneManCompany (OMC), an open-source framework that treats AI organisation design as a first-class concern.  OMC introduces three core concepts.  A \emph{Talent} is a portable agent identity package encompassing role, prompts, skills, tools, and working principles that can be deployed on any supported runtime without modification.  A \emph{Container} is the execution environment that hosts a Talent, abstracting over heterogeneous backends (LangGraph, Claude Code, script processes) through a uniform set of organisational interfaces.  Together, Talent and Container compose an \emph{Employee}, a fully managed AI agent with structured lifecycle, from hiring through the \emph{Talent Market} to performance evaluation and potential offboarding.  Mirroring how a real company operates  (as shown in Figure~\ref{fig:overview}), OMC is built on three pillars, each corresponding to a core organisational function.

The first pillar is a typed Talent-Container architecture (Section~\ref{sec:harness}) that separates \emph{who an agent is} (the Talent: prompts, skills, tools) from \emph{where it runs} (the Container: LangGraph, Claude CLI, or script process), with six typed organisational interfaces mediating all agent--platform interaction.  On the supply side, a community-driven Talent Market provides verified agent implementations that can be recruited on demand and provisioned through an automated hiring pipeline (Section~\ref{sec:talent-market}), mirroring how a human company manages its workforce under uniform HR policies while recruiting specialists from an external labour market.  Table~\ref{tab:skill-vs-talent} contrasts the Talent abstraction with conventional skills, highlighting how Talents elevate the unit of reuse from individual tools to complete agent identities with managed lifecycles.

The second pillar is an Explore-Execute-Review ($\text{E}^2$R) tree search (Section~\ref{sec:explore-review}) that models project execution as a search over organisational strategies.  The system explores decomposition options, expands the task tree, executes work through agents, and reviews results to refine future decisions.  A DAG-based task decomposition and execution mechanism (Section~\ref{sec:dag}) with AND-tree semantics and a finite state machine provides formal guarantees on termination, deadlock freedom, and crash recovery, corresponding to how a human company decomposes goals into tasks, assigns owners, reviews deliverables, and iterates until quality is met.

The third pillar is agent and organisation self-evolution (Section~\ref{sec:self-evolution}).  Agents refine their working principles through CEO one-on-ones and post-task reflection, project retrospectives distil lessons into updated Standard Operating Procedures (SOPs), and a formal HR pipeline (periodic evaluations, Performance Improvement Plans, and automated offboarding) creates real consequences, paralleling how human organisations improve through feedback, post-mortems, and performance management.

These three pillars are realised in a fully integrated system (Figure~\ref{fig:system-ui}), where the CEO manages the AI organisation through a unified interface exposing task decomposition, agent coordination, organisational knowledge, and talent lifecycle management.  To quantitatively validate this approach, we evaluate OMC on PRDBench~\cite{fu2025prdbench}, a project-level software development benchmark.  Under a single-attempt zero-shot setting, OMC achieves an 84.67\% success rate, surpassing all baselines by at least 15 percentage points.

The remainder of this paper is organised as follows.  Section~\ref{sec:methodology} presents the framework design, covering the organisational layer, $\text{E}^2$R tree search, DAG-based task execution, and self-evolution mechanisms.  Section~\ref{sec:experiments} reports quantitative results on PRDBench and four cross-domain case studies.  Section~\ref{sec:related} surveys related work.  Section~\ref{sec:discussion} discusses limitations and broader implications, and Section~\ref{sec:conclusion} concludes.

\begin{table}[t]
\centering
\small
\renewcommand{\arraystretch}{1.4}%
\rowcolors{2}{black!4}{white}
\begin{tabular}{@{}p{2.8cm}p{4.6cm}p{5.6cm}@{}}
\toprule
\textbf{Aspect} & \textbf{Skills \& Skill Markets} & \textbf{Talents \& Talent Markets\tablefootnote{\url{https://one-man-company.com/market}}} \\
\midrule
\textbf{Level}
& Inside one agent
& Across a team of agents \\
\textbf{What it is}
& Small reusable tools\newline or functions
& Full agents with roles, tools,\newline and behaviour \\
\textbf{Purpose}
& Make one agent more capable
& Build and run a team to solve tasks \\
\textbf{How they combine}
& Linked as tool chains\newline inside an agent
& Organised into teams with roles\newline and responsibilities \\
\textbf{Runtime}
& Tied to one system\newline or framework
& Can run across different systems \\
\textbf{Flexibility}
& Usually fixed\newline before execution
& Can be added, replaced,\newline or reconfigured on the fly \\
\textbf{Market role}
& Library of tools\newline to download and use
& Hiring pool of agents\newline to recruit from \\
\textbf{Quality control}
& Based on examples\newline or documentation
& Tested and evaluated agents \\
\textbf{Lifecycle}
& No clear lifecycle
& Managed lifecycle\newline (hire, evaluate, replace) \\
\textbf{Learning}
& Improves individual\newline agent performance
& Improves both agents\newline and the whole organisation \\
\textbf{Analogy}
& Software libraries (APIs)
& Employees and job markets \\
\bottomrule
\end{tabular}
\rowcolors{0}{}{}%
\renewcommand{\arraystretch}{1.0}%
\caption{Skills improve what a single agent can do. Talents organise multiple agents into a workforce that can be built, managed, and improved over time.}
\label{tab:skill-vs-talent}
\end{table}


\section{Methodology}
\label{sec:methodology}

OMC is a framework for constructing, managing, and evolving multi-agent organisations, rather than a fixed agent with a predefined workflow.  The design mirrors how a real company operates, organised around three pillars: agents are managed and recruited under uniform policies through the organisational layer and its integrated Talent Market (Section~\ref{sec:harness}), they collaborate through structured project execution via the $\text{E}^2$R tree search and DAG scheduling (Sections~\ref{sec:explore-review}--\ref{sec:dag}), and they improve over time through reflection and performance review (Section~\ref{sec:self-evolution}).

We first define the core terminology, which mirrors real-world company structures.  The fundamental unit of agency in OMC is the \textit{Employee}, analogous to an employee in a real company and characterised by domain expertise, role responsibilities, skill set, and tool access.  Each employee is decomposed into two components: a \textit{Talent}, a portable package defining the employee's cognitive identity (prompts, role and work principles, agent family configuration, tools, skills, and supporting resources, deployable across OMC instances without execution-specific dependencies), and a \textit{Container}, the runtime environment that hosts a Talent (encapsulating the agent runtime, middleware hooks, wrappers, and resources; three families are currently supported: Claude Code-based, LangGraph-based, and script-based).  The \textit{Talent Market} is a community-driven agent marketplace hosting verified and popular open-source agents; unlike systems that rely on AI-generated characters, OMC grounds agent selection in community-validated implementations, and when community-contributed agents do not cover a required domain, an AI-powered recommendation engine can discover and assemble suitable skills from the web, mitigating the cold-start problem.  The \textit{CEO} is the only human in this AI company, and also the creator and maintainer of an OMC instance; the users of OMC can be internal (CEO) or external (accessing via an AI economy contract protocol).  Each OMC instance is bootstrapped with a \textit{Founding Team} of default employees: a Human Resource Manager (HR), an Executive Assistant (EA), a Chief Operating Officer (COO), and a Chief Sales Officer (CSO).  Finally, we define a \textit{Wild Dynamic Agentic Workflow} as a multi-agent workflow in which neither the team (composition, agent runtimes, agent capabilities) nor the workflow itself (task decomposition, execution ordering) is fixed before execution; all may change at any point during a project, addressing a limitation of existing dynamic agentic workflows that adapt task decomposition at runtime but still require a pre-configured team, a single shared runtime, and a known workflow template.

\subsection{On-demand Organisation of Heterogeneous Agents by Talent--Container Architecture}
\label{sec:harness}

A core challenge is how to organise heterogeneous agents under a unified execution model.  The employees that execute tasks in OMC are radically heterogeneous: some are hosted LLM agents (LangChain/LangGraph), others are interactive coding sessions (Claude Code), and still others are script-based executors wrapping open-source frameworks.  Without a unifying abstraction, the orchestration layer would degenerate into a tangle of backend-specific conditional logic, and every new agent runtime would require invasive changes across scheduling, lifecycle management, and event propagation.  OMC solves this through a typed organisational layer that standardises how any agent backend connects to the platform, analogous to how an OS kernel provides a uniform interface over heterogeneous hardware.  Together, the Talent and Container abstractions form a \emph{digital talent layer}: a composable, runtime-independent representation of agent capabilities that sits between low-level skills and high-level organisational structure.

\subsubsection{Employee: Talent and Container Composition}

\begin{figure}[ht]
\centering
\includegraphics[width=\textwidth]{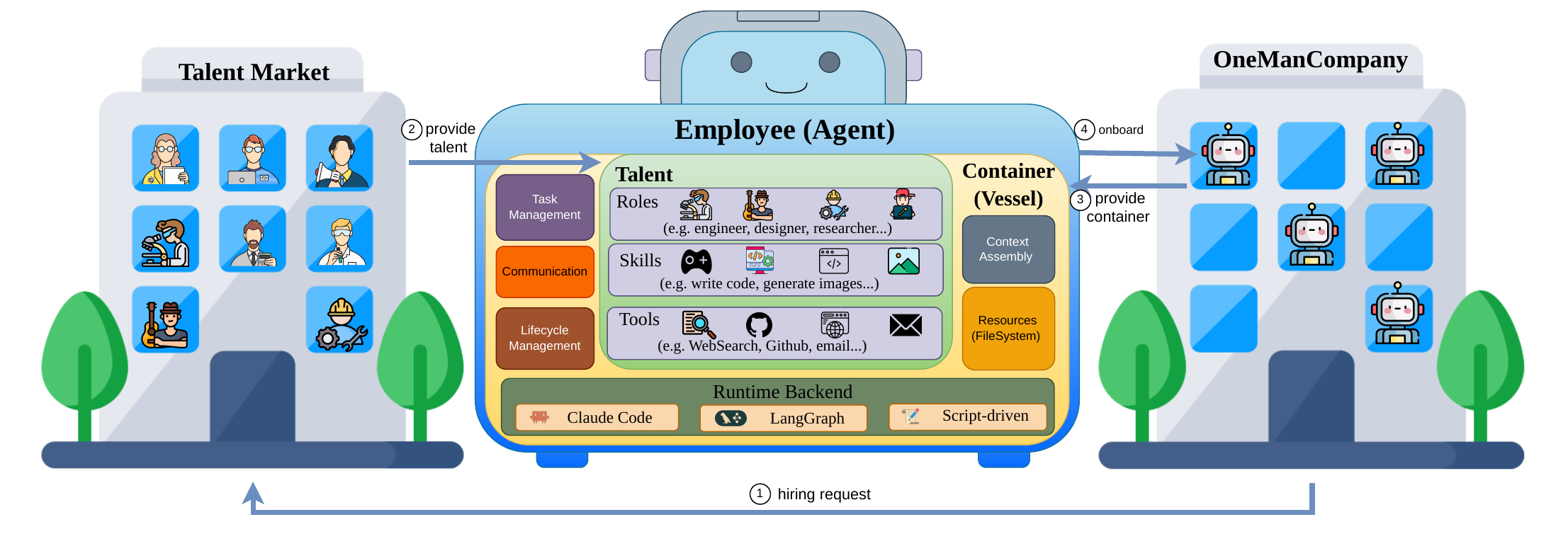}
\caption{\textbf{Employee = Talent + Container.}  Each employee in OMC is composed of a \textit{Talent} (a portable agent package encapsulating role, skills, and tools) which hire from the Talent Market  , and a \textit{Container} (a runtime backend such as LangGraph, Claude Code, or script-driven, together with six organisational interfaces: execution, task management, event communication, storage, context assembly, and lifecycle management).  The same Talent can be deployed on any supported Container, enabling heterogeneous agents to coexist in a single organisation.}
\label{fig:employee-composition}
\end{figure}

As illustrated in Figure~\ref{fig:employee-composition}, each employee is composed of a \textit{Talent} (portable cognitive identity) and a \textit{Container} (execution runtime plus organisational interfaces).  The Container not only hosts the agent runtime but also provides the organisational layer, i.e.\ the formal contract through which it exposes its capabilities to the OMC platform.  It decomposes agent--platform interaction into six typed organisational interfaces: \emph{Execution} dispatches a task to the backend and returns the result; \emph{Task} manages a per-employee queue with mutual exclusion; \emph{Event} provides an organisational event bus for publish/subscribe communication; \emph{Storage} handles persistent memory for short- and medium-term state; \emph{Context} assembles the execution prompt from the Talent's role, guidance, and memory; and \emph{Lifecycle} applies pre- and post-execution hooks for validation, guardrails, and self-improvement.  Formal signatures are provided in Appendix~\ref{sec:harness-signatures}; Algorithm~\ref{alg:harness} illustrates how the interfaces compose during task execution (conventions: \textit{italic} = variables, \textsc{SmallCaps} = functions, $x.\mathrm{field}$ = field access).

This design yields three properties.  (1)~\emph{Identity--substrate separation}: the same Talent can run on a LangGraph agent, a Claude CLI session, or a script-based executor without modification, and conversely the same Container type can host different Talents to produce employees with different roles on the same backend.  (2)~\emph{Multi-tenancy with isolation}: all agent--platform interaction passes through the six typed interfaces, so no Container can bypass organisational policies on task validation, event propagation, or memory access.  (3)~\emph{Extensibility without platform modification}: adding a new agent runtime requires only a Container implementation conforming to the six contracts, and the lifecycle hooks enable self-improvement (working principle refinement, skill accumulation) entirely at the organisational layer, without modifying underlying foundation models.  We observe that the six interfaces mirror the canonical subsystems of an OS kernel (process management, memory, file system, I/O, IPC, security)~\cite{tanenbaum2014modern, silberschatz2018os}; the detailed correspondence is provided in Appendix~\ref{sec:os-mapping}.

\begin{algorithm}[t]
\caption{Talent Assembly.}
\label{alg:harness}
\begin{algorithmic}[1]
\Require employee $e$ with Container $V_e$, Talent $\tau_e$, and task node $v$
\Ensure result $r$, cost $c$
\State \textbf{// Queue management: enforce mutual exclusion}
\If{$e$ has a running task}
    \State \textsc{Enqueue}($v$) and \Return
\EndIf
\State \textbf{// Context assembly: build execution prompt}
\State $\mathit{ctx} \gets \textsc{AssembleContext}(\tau_e.\mathrm{role},\; \tau_e.\mathrm{principles},\; \textsc{GetGuidance}(e),\; \textsc{GetMemory}(e))$
\State \textbf{// Lifecycle pre-hook: validation and enrichment}
\State $v,\; \mathit{ctx} \gets \textsc{PreHook}(v,\; \mathit{ctx})$ \Comment{guardrails, input validation}
\State \textbf{// Dispatch to backend via Container}
\State $r,\; c \gets V_e.\textsc{Execute}(v.\mathrm{desc},\; \mathit{ctx},\; \tau_e.\mathrm{tools})$
\State \textbf{// Lifecycle post-hook: memory update and skill refinement}
\State \textsc{PostHook}($e,\; v,\; r$) \Comment{self-reflection, principle updates}
\State \textbf{// Event publication}
\State \textsc{Publish}($\varepsilon(v,\; \textsc{Processing} \to \textsc{Completed},\; e,\; t_{\mathrm{now}})$)
\State \Return $r,\; c$
\end{algorithmic}
\end{algorithm}

\subsubsection{Digital Talent Market}
\label{sec:talent-market}

OMC integrates a community-driven Talent Market as a native capability layer.  Rather than synthesising agents from descriptive prompts, a practice prone to capability hallucination, OMC recruits from a pool of community-verified, benchmark-validated implementations and provisions them through an automated hiring pipeline.

Each Talent on the marketplace is a complete, ready-to-deploy agent package comprising system prompts and role definitions, tool configurations and MCP integrations, skill scripts, domain knowledge files, and benchmark results that document the agent's verified capabilities.  Talents span a wide range of specialisations (software engineers, data analysts, content writers, art designers, QA testers, research analysts, among others), each grounded in executable artefacts rather than descriptive claims.  Because Talents are decoupled from Containers (Section~\ref{sec:harness}), the same Talent can be deployed on any supported runtime (LangGraph, Claude CLI, script-based), and the same OMC instance can recruit Talents built on entirely different agent families to form a heterogeneous team.  The marketplace supports three sourcing channels: (1)~\emph{community-contributed Talents}, open-source agent packages uploaded and peer-reviewed by the community; (2)~\emph{AI-recommended assembly}, an AI-powered engine that discovers suitable skills and tools from the web and assembles them into functional Talent packages, mitigating the cold-start problem for underserved domains; and (3)~\emph{internal promotion}, high-performing employees whose refined profiles and accumulated skills are packaged and shared back to the marketplace.

When the policy $\pi(\mathcal{T})$ requires a capability absent from the current workforce, triggering a recruit action $\alpha_{\text{r}} \in \mathcal{A}_{\text{recruit}}$, the HR agent queries the Talent Market, compiles a ranked shortlist based on skill match and community ratings, and presents candidates to the CEO for approval.  Upon selection, the automated pipeline provisions the Talent with a Container, assigns a desk, configures tool access, and registers the new employee in the organisational hierarchy, all without manual setup.  This on-demand recruitment closes the loop between capability demand (what the project needs) and supply (what the community has built), enabling OMC to tackle projects across arbitrary domains without pre-configuring a fixed team.

\subsection{\texorpdfstring{Strategy Search in Vast Organisational Space by $\text{E}^2$R Tree}{Strategy Search in Vast Organisational Space by E²R Tree}}
\label{sec:explore-review}

Consider a company that receives a project request. The space of possible responses is enormous: the project can be decomposed in multiple ways, each subtask can be assigned to different employees, each intermediate result can be accepted, rejected, or iterated upon, and the entire decomposition can be revised after observing partial outcomes. Existing multi-agent frameworks typically handle this by either hardcoding a fixed workflow graph or allowing agents to freely negotiate, neither of which scales: fixed workflows cannot adapt to novel projects, while unconstrained negotiation provides no convergence guarantees.

We observe that this problem shares key structural properties with game-tree search: a large branching factor, stochastic outcomes from non-deterministic LLM execution, and the need to balance exploration and exploitation. Drawing on the structural principles of Monte Carlo Tree Search (MCTS)~\cite{kocsis2006bandit}, OMC decomposes the organisational decision cycle into an \textit{Explore-Execute-Review ($\text{E}^2$R)}: \emph{explore} the strategy space (select a decomposition and expand the task tree), \emph{execute} the plan (agents carry out assigned work), and \emph{review} the results (propagate quality signals to refine future decisions).  Unlike MCTS, E$^2$R does not use simulated rollouts or UCB-based selection; execution is real (agents produce actual deliverables, not value estimates), and the review signal comes from explicit supervisor evaluation rather than terminal reward backpropagation.  The analogy is structural: both methods grow a tree incrementally, evaluate nodes after expansion, and use the evaluation to guide subsequent exploration.

\subsubsection{Tree Structure: Nodes and Edges}

The $\text{E}^2$R operates over a \emph{search tree} $\mathcal{T} = (V, E_{\text{tree}}, E_{\text{dep}})$ that grows dynamically during project execution.

\paragraph{Nodes.}
Each node $v \in V$ represents the organisational state at a particular decision point, comprising task-level attributes and shared organisational context.  Each $v$ carries:
\begin{equation}
(d_v,\; e_v,\; \phi_v,\; r_v,\; c_v,\; \mathcal{W},\; \mathcal{R})
\end{equation}
where $d_v$ is the task description, $e_v \in W \cup \{\varnothing\}$ is the assigned employee drawn from the current workforce $W$ (or $\varnothing$ if unassigned), $\phi_v \in \Phi$ is the current status from the finite state machine defined in Section~\ref{sec:dag}, $r_v$ is the result produced upon completion (initially $\varnothing$), and $c_v \in \mathbb{R}_{\geq 0}$ is the accumulated execution cost.  The shared components $\mathcal{W}$ (workforce state: the set of employees $W$ together with their skills, workload, and performance history) and $\mathcal{R}$ (resource state: token budget, cost accumulation, time constraints) are tree-wide and referenced by all nodes.  A single project directive from the CEO creates the \textit{root node}; from that point onward, the tree grows through decisions made by the agent responsible for each node, producing \textit{interior nodes} (sub-tasks decomposed by a parent's owner) and \textit{leaf nodes} (atomic tasks assigned to and executed by individual employees).

\paragraph{Edges.}
Two types of edges structure the tree.  \textit{Decomposition edges} $E_{\text{tree}} \subseteq V \times V$ encode strategy: a directed edge $(p, v')$ means ``parent $p$ was decomposed into child $v'$''; these form a strict tree (each child has exactly one parent), and different decomposition choices produce different subtrees.  \textit{Dependency edges} $E_{\text{dep}} \subseteq V \times V$ encode execution ordering: a directed edge $(u, v)$ means ``$v$ cannot begin execution until $u$ is accepted''; these constraints operate \emph{within} a decomposition strategy and may cross sibling branches (e.g., the frontend task depends on the API task, even though both are children of the same parent).  The combined graph $G = (V, E_{\text{tree}} \cup E_{\text{dep}})$ must be a DAG, enforced at insertion time via DFS cycle detection.

\paragraph{Actions.}
At each decision point, the system selects from five action types that modify the tree:
\begin{equation}
\mathcal{A} = \mathcal{A}_{\text{decompose}} \cup \mathcal{A}_{\text{assign}} \cup \mathcal{A}_{\text{recruit}} \cup \mathcal{A}_{\text{review}} \cup \mathcal{A}_{\text{iterate}}
\end{equation}
$\mathcal{A}_{\text{decompose}}$ adds decomposition edges (new children under a node); $\mathcal{A}_{\text{assign}}$ binds an employee to a leaf node; $\mathcal{A}_{\text{recruit}}$ hires a new employee from the Talent Market when required capabilities are missing (Section~\ref{sec:talent-market}); $\mathcal{A}_{\text{review}}$ transitions a node's status (accept or reject); and $\mathcal{A}_{\text{iterate}}$ creates a new root-level iteration with an updated strategy.

\paragraph{Strategies and Policy.}
A \emph{strategy} $\sigma$ is a sequence of actions applied to the current tree.  The structural transition function $T$ maps a tree--action pair to a successor tree deterministically (adding nodes, updating edges, changing statuses); stochasticity enters during the Execute stage (Stage~2 below), where each agent's internal reasoning is non-deterministic:
\begin{equation}
\sigma = (a_1, a_2, \ldots, a_m), \quad a_i \in \mathcal{A}, \quad \mathcal{T}_{i+1} = T(\mathcal{T}_i, a_i)
\end{equation}
In the Explore stage, the system must select a strategy for the current decision point.  We define a \emph{policy} $\pi$ that maps a tree to a strategy (a complete decomposition-and-assignment plan for the current decision point):
\begin{equation}
\pi(\mathcal{T}) = \sigma = \bigl(\alpha_{\text{d}}(v, \{v'_1, \ldots, v'_n\}),\; \alpha_{\text{a}}(v'_1, e_1),\; \ldots,\; \alpha_{\text{a}}(v'_n, e_n)\bigr)
\end{equation}
where $v$ is the node to be expanded, $\{v'_1, \ldots, v'_n\}$ are the new child tasks, $\alpha_{\text{d}}$ denotes a decompose action, $\alpha_{\text{a}}$ denotes an assign action, and $e_i$ is the employee assigned to $v'_i$ (possibly via a recruit action if no suitable employee exists).  The composite operation $\Delta(v, e, d, D)$ applies one decompose--assign pair, creating a child node with description $d$, assigning employee $e$, and registering dependency edges $D \subseteq V \times V$:
\begin{equation}
\quad v_{\text{new}} = \Delta(v, e, d, D)
\end{equation}
In the current implementation, $\pi$ is realised by the supervising agent (e.g., the COO or a senior employee) reasoning over the project state, employee profiles, and accumulated performance history.  The supervisor serves as the heuristic policy that jointly decides decomposition granularity and employee assignment.

\begin{wrapfigure}{r}{0.4\textwidth}
    \centering
    \vspace{-12pt}
    \includegraphics[width=0.38\textwidth]{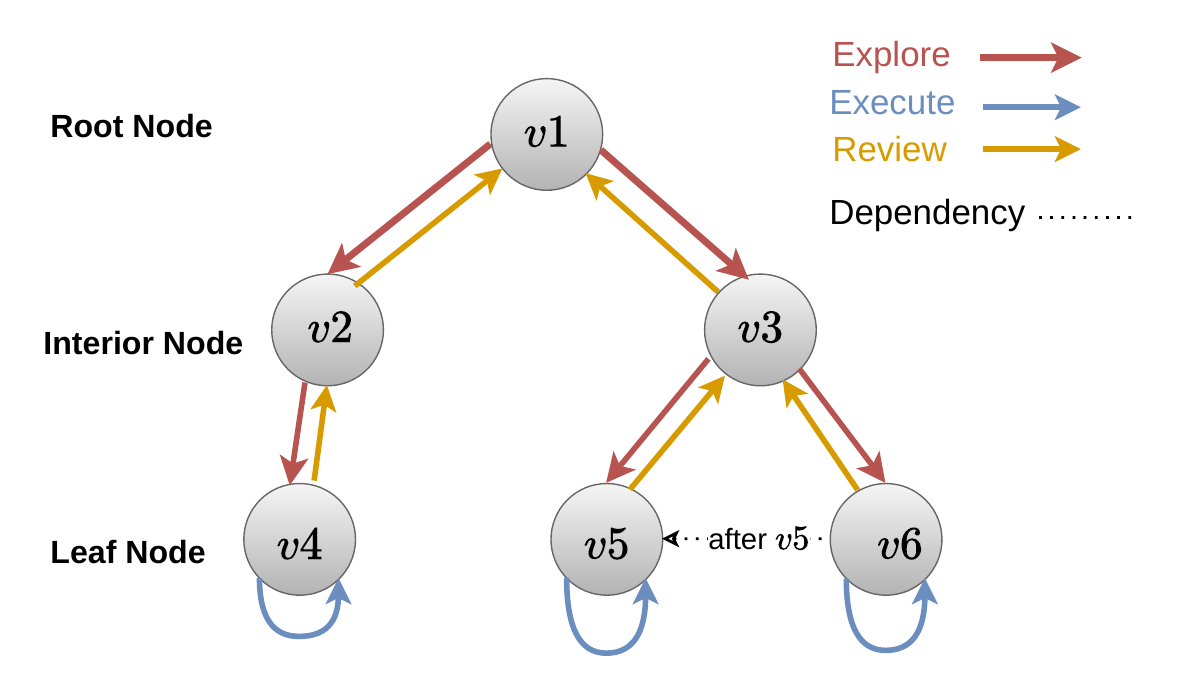}
    \caption{An illustration of the $\text{E}^2$R tree search loop: Explore, Execute, Review.}
    \label{fig:e2r-loop}
\end{wrapfigure}

\paragraph{Three Stages of Explore-Execute-Review.} Each iteration of the search loop applies $\pi$, executes the resulting strategy, and feeds back quality signals, in three stages (Figure~\ref{fig:e2r-loop}):

\textit{Stage 1: Explore} (strategy selection and task tree expansion).
The executive agents apply $\pi(\mathcal{T})$ to select a strategy: how to decompose the current task and whom to assign.  This stage faces the classic exploration--exploitation trade-off: assign tasks to employees with proven track records (exploitation) or try less-tested employees, or hire a new one, to discover hidden capability (exploration).  The branching factor is unbounded, since the LLM decides decomposition granularity at runtime.

\textit{Stage 2: Execute} (agents carry out assigned work).
Each assigned employee executes its task through the organisational layer (Section~\ref{sec:harness}).  We write $f_{e_v}$ for the internal execution function of employee $e_v$, which takes the task description $d_v$ and produces a result and cost:
\begin{equation}
(r_v, c_v) = f_{e_v}(d_v)
\end{equation}
where $r_v$ is the result and $c_v$ is the execution cost.  The internal function $f$ is determined by the agent's own loop (e.g., reasoning, tool use, code generation) and, for closed-source agents such as Claude, is opaque to the organisational layer.  The quality signal $q_v$ is produced separately in Stage~3 when a reviewer evaluates $r_v$.  The DAG execution layer (Section~\ref{sec:dag}) formalises the dependency resolution and termination guarantees for this stage.

\textit{Stage 3: Review} (quality signals propagate and drive iteration).
For each completed node, a reviewer (typically the node's parent owner or the COO) evaluates the result $r_v$ and produces a quality signal $q_v \in \{\textsc{accept}, \textsc{reject}\}$, which triggers the corresponding status transition in the FSM.  These review decisions propagate bottom-up from leaf nodes to the root, accumulating into a review signal vector $\mathbf{g}(v)$ per node:
\begin{equation}
\mathbf{g}(v) = \bigl(q_v,\; c_v,\; \phi_v\bigr), \quad \forall v \in \text{path}(\text{leaf}, \text{root})
\end{equation}

A subtree reaches its bottom when every leaf node has received an assign action $\alpha_{\text{a}}$ and been executed; at that point, the review action $\alpha_{\text{rev}}$ determines whether each task has been completed satisfactorily.  If a reviewer accepts the result, the quality signal propagates upward and may unblock dependent nodes or resolve the parent.  If a reviewer rejects the result, the system re-enters Stage~1: a new decomposition is explored under the same parent, effectively growing a fresh subtree with updated context from the failed attempt.  This accept-or-redecompose cycle continues until the root is resolved or a circuit breaker fires (Section~\ref{subsubsection:brcb}).  Whether a node counts as resolved is defined by the AND-semantics in Section~\ref{sec:dag}.  Accumulated signals are persisted in task histories and employee progress logs, enriching the organisational context for future iterations.

\subsubsection{Iterated Search with External Oracle}

Each project iteration ($\texttt{iter\_001}, \texttt{iter\_002}, \ldots$) corresponds to one search episode.  We define the policy update function $\Pi$ that refines the organisational policy after each iteration.  In the current implementation $\pi$ is not a parameterised policy updated by gradient methods; rather, $\Pi$ enriches the context available to the supervising agent (accumulated history, updated employee profiles, revised SOPs) so that subsequent calls to $\pi$ produce better strategies:
\begin{equation}
\pi_{k+1} = \Pi(\mathcal{T}_k, \pi_k, \mathcal{H}_k)
\end{equation}
where $\mathcal{H}_k = \{(v, r_v, c_v, q_v)\}_{v \in \text{completed}}$ is the accumulated execution history (results, costs, and review signals from all completed nodes up to iteration $k$).

The CEO (or an external customer of the company) acts as an \emph{external oracle} who provides three types of intervention: (1) \emph{policy override}, directly rejecting or redirecting a decomposition strategy; (2) \emph{requirement injection}, adding new constraints mid-search (``add SEO support'', ``change the architecture''); and (3) \emph{iteration triggering}, deciding when to launch a new search episode and when to stop. We model this human stakeholder as a meta-level controller~\cite{russell1991right} who applies domain-informed optimal stopping, iterating when the expected improvement justifies the cost and stopping when marginal returns diminish.

This human-in-the-loop design has a practical advantage: the stakeholder prunes unpromising branches early, injects external information unavailable to the system (market signals, strategic priorities), and concentrates computation on high-value regions of the search space. The trade-off is that convergence depends on the quality of human judgment rather than on formal guarantees.

\subsubsection{Bounded Rationality and Circuit Breakers}
\label{subsubsection:brcb}
Real organisations cannot run infinite simulations.  OMC implements bounded rationality through three mechanisms, where $n_{\text{rev}}(v)$ denotes the number of review rounds for node $v$, $t_{\text{exec}}(v)$ its wall-clock execution time, and $\text{esc}(v)$ the escalation of $v$ to a higher-level supervisor: a \emph{review round limit} ($n_{\text{rev}}(v) \geq k_{\text{rev}} \implies \text{esc}(v)$, default $k_{\text{rev}} = 3$), a \emph{task timeout} ($t_{\text{exec}}(v) > T_{\max} \implies \phi_v \leftarrow \textsc{failed}$, default $T_{\max} = 3600$s), and a \emph{cost budget} ($\sum_{v \in V} c_v > B \implies \text{pause}$).  All three are configurable.  Together, these mechanisms guarantee that every search episode terminates in bounded time and cost under the assumption that the underlying executor (LLM, tool calls, external services) respects the timeout contract.  Section~\ref{sec:dag} formalises the stronger execution guarantees that hold within these bounds.

\subsubsection{DAG-based Task Decomposition and Execution}
\label{sec:dag}

The $\text{E}^2$R tree search and DAG execution form two complementary layers: each search \emph{iteration} corresponds to one complete DAG execution cycle, and the DAG layer returns the signals (quality, cost, time) that drive the review phase.  We write $\mathcal{I}$ (for \emph{iteration}) to denote this combined function:
\begin{equation}
(\bar{r}, \bar{c}, \bar{q}) = \mathcal{I}(\mathcal{T}, a) = \mathcal{S}_{\text{DAG}}(T(\mathcal{T}, a))
\end{equation}
where $T(\mathcal{T}, a)$ is the successor tree produced by applying decomposition action $a$ to tree $\mathcal{T}$, $\mathcal{S}_{\text{DAG}}$ is the DAG scheduler that executes all ready nodes, resolves dependencies, and propagates review signals (defined by the scheduling, propagation, and guarantee rules below), and $(\bar{r}, \bar{c}, \bar{q})$ are the aggregated result, cost, and quality signals returned to the $\text{E}^2$R review phase.

Once the exploration phase selects a decomposition strategy, the resulting task graph must be executed reliably.  This raises a concrete question: how can the system guarantee task completion when the decomposition itself is dynamic?  The challenge is non-trivial: tasks may depend on each other in arbitrary DAG patterns, employees may fail or produce unacceptable results, and the system must recover from crashes without losing progress. Without formal guarantees, a multi-agent system risks silent stalls: tasks that never reach a terminal state, dependencies that permanently block downstream work, or review loops that cycle indefinitely.

We formalise task execution as scheduling over an AND-tree augmented with dependency edges, where a finite state machine governs each node's lifecycle with termination guarantees under bounded retry and finite resource constraints.
\begin{figure}[t]
    \centering
    \includegraphics[width=\textwidth]{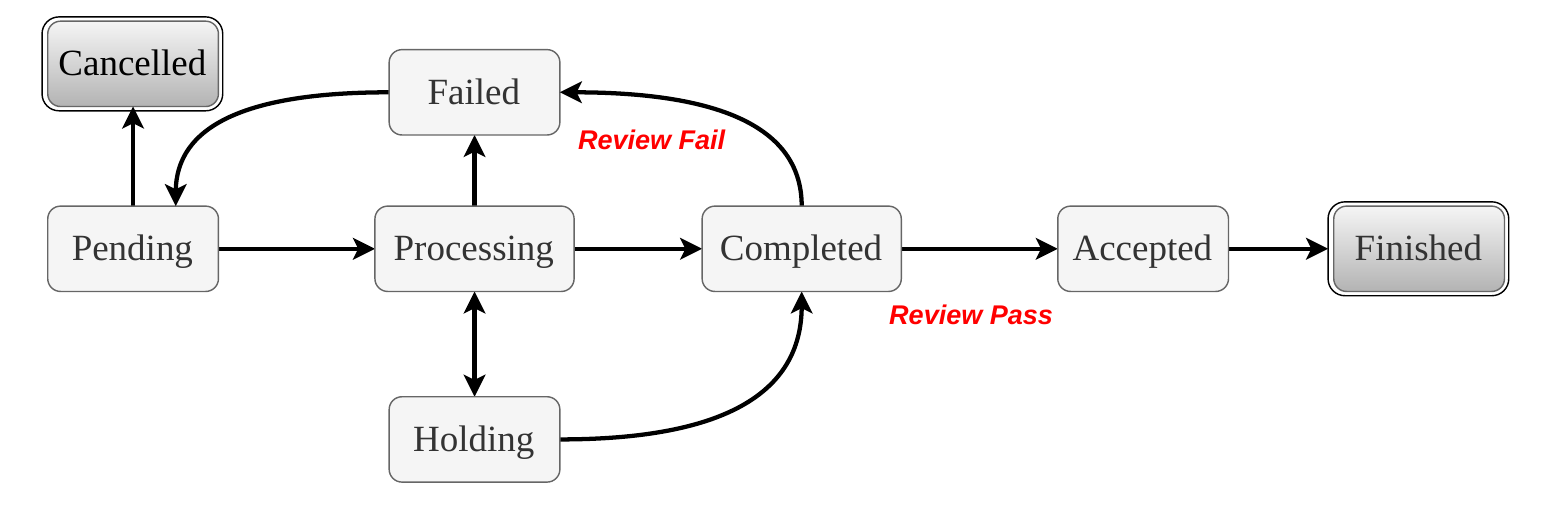}
\caption{\textbf{Task lifecycle finite state machine (FSM).} Double-bordered states are terminal. The \textsc{completed}$\to$\textsc{accepted} transition requires explicit supervisor review, preventing unverified results from propagating downstream.}
\label{fig:task-states}
\end{figure}
\paragraph{AND-Tree with DAG Dependencies.} The task tree $\mathcal{T} = (V, E_{\text{tree}}, E_{\text{dep}})$ and its DAG invariant were defined above.  Here we specify the execution semantics that govern how this tree is scheduled and completed.

\paragraph{AND-Semantics.}
A node $v$ is \emph{resolved} according to the following recursive definition:
\begin{equation}
\text{resolved}(v) \iff \begin{cases} \phi_v \in \{\textsc{accepted}, \textsc{finished}\} & \text{if } v \text{ is a leaf} \\ \forall v' \in \text{children}(v) \setminus S : \text{resolved}(v') & \text{otherwise} \end{cases}
\end{equation}
where $S$ is the set of system node types (review requests, watchdog nudges).  This AND-semantics is the key structural guarantee: completion \emph{must} propagate bottom-up from leaves through the entire tree. No subtasks can be silently dropped.

\paragraph{Task Lifecycle State Machine.} Each task node follows a finite state machine $M = (\Phi, \delta, \phi_0, F)$ (Figure~\ref{fig:task-states}) with initial state $\phi_0 = \textsc{pending}$, terminal states $F = \{\textsc{finished}, \textsc{cancelled}\}$, and state set:
\begin{equation}
\Phi = \left\{\;\begin{aligned}
& \textsc{pending},\; \textsc{processing},\; \textsc{holding},\\
& \textsc{completed},\; \textsc{accepted},\; \textsc{failed},\\
& \textsc{blocked},\; \textsc{finished},\; \textsc{cancelled}
\end{aligned}\;\right\}
\end{equation}

Two design choices are critical (Figure~\ref{fig:task-states}). First, \textsc{completed}$\to$\textsc{accepted} requires explicit supervisor review, which prevents hallucinated or incorrect results from unblocking dependent tasks. Second, the \textsc{failed}$\to$\textsc{processing} retry path is bounded by a maximum retry count $k_{\text{retry}}$; exceeding this limit triggers escalation, guaranteeing that no node cycles indefinitely.

\paragraph{Scheduling and Dependency Resolution.} A node $v$ becomes executable when its dependency constraints are satisfied:
\begin{equation}
\text{ready}(v) \iff \phi_v = \textsc{pending} \;\wedge\; \forall u \in \text{deps}(v) : \phi_u \in \{\textsc{accepted}, \textsc{finished}\}
\end{equation}

The scheduler selects the first ready node per employee in FIFO order, subject to a mutual exclusion invariant: $|\text{running}(e)| \leq 1$ for all employees $e$.  When a node reaches a resolved state, dependency resolution propagates forward through the DAG: dependent nodes with all dependencies resolved are scheduled for execution; dependent nodes with failed dependencies are marked \textsc{blocked} (or cascade-cancelled if the failure is non-recoverable); and cascade cancellation implements transitive closure ($\text{cancel}(v) \implies \forall w : v \in \text{deps}(w) \implies \text{cancel}(w)$).

\paragraph{Bottom-Up Completion Propagation.}

When a leaf node completes and passes review, the AND-semantics trigger recursive propagation: if all children of a parent are resolved, the parent auto-promotes to \textsc{completed}$\to$\textsc{accepted}$\to$\textsc{finished}, which in turn triggers its own parent's resolution check, and so on up to the project root. This bottom-up propagation ensures that project completion is a \emph{derived} property of subtask completion, not a separately maintained flag.

A deadlock detector provides a safety net: if all non-root nodes are in terminal or blocked states but the root has not resolved, the project is marked as failed, preventing silent stalls.

The combined design provides seven invariants: (1)~\emph{DAG Invariant}: $G = (V, E_{\text{tree}} \cup E_{\text{dep}})$ is always acyclic, enforced at insertion; (2)~\emph{Mutual Exclusion}: $|\text{running}(e)| \leq 1$ for all employees $e$; (3)~\emph{Schedule Idempotency}: repeated scheduling of the same node is a no-op, so crash recovery never causes duplicate execution; (4)~\emph{Review Termination}: at most $k_{\text{rev}}$ reviews per parent before escalation; (5)~\emph{Cascade Completeness}: cancellation propagates to all transitive dependents; (6)~\emph{Dependency Completeness}: every resolved-state transition triggers forward dependency resolution, so no dependent is left permanently \textsc{pending}; and (7)~\emph{Recovery Correctness}: after crash, \textsc{processing} nodes reset to \textsc{pending} and all nodes with resolved dependencies are re-scheduled, ensuring the system resumes from a consistent state.

\subsection{Self-Evolution: How Agents and the Organisation Improve}
\label{sec:self-evolution}

The components described so far handle execution, but not learning.  In a real company, employees grow through one-on-ones with their manager, projects end with post-mortems, and periodic performance reviews create accountability.  OMC implements all three.

\subsubsection{Individual-Level Evolution}

Each agent maintains a persistent, auto-updating profile comprising a cross-task progress log and LLM-summarised working principles.  Two triggers drive individual reflection.  First, after each \emph{CEO one-on-one}, the agent performs structured self-reflection: reviewing the CEO's feedback, identifying gaps between expected and actual behaviour, and updating its working principles accordingly, mirroring the manager--report one-on-one in human organisations.  Second, upon task completion, the agent conducts a \emph{post-task review} of its own execution trace (decisions made, tools invoked, obstacles encountered) and appends a summary to its progress log; over successive tasks, this log accumulates a trajectory of lessons learned that enriches the agent's context for future work.

Critically, these updates modify the agent's Talent artefacts (working principles, guidance notes) rather than the underlying foundation model, enabling continuous improvement without retraining.  In the formal model, updated working principles are reflected in the workforce state $\mathcal{W}$, so subsequent calls to $\pi(\mathcal{T})$ see the improved agent profiles.

\subsubsection{Organisation-Level Evolution}

At the project level, OMC conducts structured retrospectives that mirror real-world project post-mortems.  When a project reaches completion, the COO convenes a retrospective session in which each participating employee submits a self-assessment summarising key decisions, obstacles encountered, and solutions adopted.  The COO aggregates these self-assessments with objective signals (per-task retry counts, review rejection reasons, and resource consumption) and distils the findings into two outputs: \emph{individual feedback} that updates each employee's working principles, and \emph{organisational SOPs} that codify effective patterns for future projects (e.g., ``mandate API contract review before frontend--backend integration'').  These SOPs are persisted as workflow documents and automatically injected into relevant agents' contexts in subsequent projects, ensuring that organisational knowledge accumulates across projects rather than remaining confined to individual agent memories.

\paragraph{Performance Review and HR Lifecycle.} To create accountability and prevent capability stagnation, OMC implements a formal performance review pipeline inspired by real-world HR practices.  Every three projects, the HR agent automatically initiates a \emph{periodic review} for each participating employee, assessing task completion quality, review pass rates, and collaboration effectiveness.  An employee who fails three consecutive reviews enters a formal \textit{Performance Improvement Plan (PIP)}, receiving targeted coaching, adjusted task assignments, and closer supervision.  If the employee fails one additional review under PIP, the system triggers \emph{automated offboarding}: the agent's Container is deprovisioned, its desk is freed, and the capability gap is flagged for re-recruitment from the Talent Market.

This lifecycle management closes the loop between the Talent Market (Section~\ref{sec:talent-market}) and organisational evolution: underperforming agents are replaced by fresh recruits, while high-performing agents accumulate experience that makes them increasingly effective.  We are not aware of prior work that applies structured HR protocols (performance reviews, PIP, formal offboarding) to AI agent lifecycle management.

\section{Experiments}
\label{sec:experiments}

We evaluate OMC on PRDBench~\cite{fu2025prdbench}, a recently proposed benchmark for assessing LLM-based code agents in realistic software development scenarios.  PRDBench consists of 50 project-level tasks spanning over 20 domains, where each task is defined by a structured Product Requirement Document (PRD) along with comprehensive evaluation criteria.  Each test case provides high-level requirements, auxiliary data, detailed test plans, and executable evaluation scripts, enabling end-to-end assessment of an agent's ability to interpret requirements, decompose tasks, implement solutions, and satisfy functional constraints.  Unlike traditional benchmarks that focus on isolated code generation or unit-level correctness, PRDBench evaluates agents at the project level, requiring long-horizon reasoning, hierarchical task decomposition, and coordinated multi-agent execution. Each PRDBench task constitutes a wild dynamic agentic workflow in the sense defined in Section~\ref{sec:methodology}: neither the team (composition, runtimes, capabilities) nor the workflow (task decomposition, execution ordering) is known before execution, making it particularly suitable for evaluating OMC.

\subsection{Experimental Setup}

For the agent setup, in addition to our founding agent (a LangGraph-based agent using the Gemini 2.1 Flash Lite Preview model), the HR recruited three specialised employees from the Talent Market at the outset of the first PRD project. The first is a Software Engineer (a Claude Code-based agent with the superpowers plugin\footnote{\url{https://github.com/obra/superpowers}}), the second is a Software Architect (a Claude Code-based agent from the agency-agents project\footnote{\url{https://github.com/msitarzewski/agency-agents}}), and the third is a Code Reviewer, also sourced from the agency-agents project.

We follow the official DEV mode setting, in which each system receives the PRD as a one-shot input and must complete the task without iterative feedback or external intervention, with final outputs directly evaluated by automated scripts. We compare OMC against the baseline results reported in PRDBench under the same setting, using \emph{Success Rate} as the primary metric, defined as the percentage of tasks successfully completed. In addition, we report the \emph{Cost Overhead} of our method (e.g., token usage or API cost) to reflect the efficiency of the entire agent system under the zero-shot conditions.

\subsection{Main Results}

\begin{table}[t]
\centering
\caption{Performance comparison on PRDBench.}
\label{tab:prdb_results}
\small
\begin{tabular}{lccc}
\toprule
\textbf{AgentType} & \textbf{Method} & \textbf{Success Rate (\%)} & \textbf{Cost (\$)} \\
\midrule
Minimal & GPT-5.2 & 62.49 & - \\
Minimal & Claude-4.5 & 69.19 & - \\
Minimal & Gemini-3-Pro & 22.76 & - \\
Minimal & Qwen3-Coder & 43.84 & - \\
Minimal & Kimi-K2 & 20.52 & - \\
Minimal & DeepSeek-V3.2 & 40.11 & - \\
Minimal & GLM-4.7 & 38.39 & - \\
Minimal & Minimax-M2 & 17.60 & - \\
Commercial & CodeX & 62.09 & - \\
Commercial & Claude Code & 56.65 & - \\
Commercial & Gemini CLI & 11.29 & - \\
Commercial & Qwen Code & 39.91 & - \\ \hline
\textbf{Multi-agent} & \textbf{\makecell{Ours (Claude Code Sonnet 4.6 \\+ Gemini 3.1 Flash Lite Preview)}} & \textbf{84.67 (+15.48)} & \textbf{345.59} \\
\bottomrule
\end{tabular}
\end{table}

As shown in Table~\ref{tab:prdb_results}, OMC achieves the highest success rate of 84.67\% (+15.48\%) across all systems, surpassing all baselines.  The total cost of \$345.59 across 50 tasks (approximately \$6.91 per task) reflects the overhead of multi-agent coordination; cost data for baselines was not reported in PRDBench, so a direct cost-efficiency comparison is not possible.

Three aspects of OMC's design contribute to this result.  First, the dynamic task tree adjusts decomposition during execution based on intermediate results, rather than committing to a fixed pipeline upfront.  Second, the enforced \textsc{completed}$\to$\textsc{accepted} review gate means no subtask result propagates downstream without supervisor approval, which reduces hallucinated outputs and limits error cascading.  Third, the Container--Talent separation lets the system recruit agents from different families (LangGraph, Claude CLI, script-based) within the same project, so the right tool is matched to each subtask.

\subsection{Case Study}

\subsubsection{Dynamic Team Assembly for Content Generation}
\label{sec:case-content}

We illustrate OMC's dynamic team assembly and cross-model collaboration through a
content-generation task in which the CEO instructs the system to assemble a
search-and-writing team, produce a weekly trend summary of hot AI Agent repositories
on GitHub, and email the finished article with real links to a designated address.
The entire company is triggered by a \emph{single} CEO prompt:

\begin{promptbox}
\noindent{\ttfamily\spaceskip=0.4em plus 0.2em minus 0.1em Assemble a search-and-writing team, produce a weekly trend summary article of the hottest AI Agent repositories on GitHub from the past week, and email it to me upon completion with links included, all repositories must be real. Send to joedoe@email.com.}
\end{promptbox}

\noindent The only human involvement throughout the entire execution consisted of submitting the prompt above and selecting employees from the HR-compiled candidate shortlist; all subsequent planning, coordination,
execution, and delivery were performed autonomously by the system. The assembled team, output artefacts, and cost breakdown are demonstrated in Figure~\ref{fig:case-content}. The \textbf{EA} decomposes the directive into two sequential phases and forwards
Phase~1 to \textbf{HR}, which queries the Talent Market, compiles a ranked shortlist,
and, upon CEO approval, provisions a \textbf{Researcher} (GPT-4o) and a
\textbf{Writer} (Claude Sonnet~4) with appropriate Containers.
Once the team is assembled, the \textbf{COO} constructs a dynamic task tree and
dispatches Phase~2: the Researcher collects verified GitHub repository links and
summaries into \texttt{research\_data.md}, and the Writer drafts a comprehensive
trend report and delivers it via email. The entire pipeline completes in under
10 minutes at a total cost of approximately \$4.48.

We manually verified all repository links and star counts reported in the
autonomously generated article, confirming that every entry is real and accurate. The final article delivered to the CEO is reproduced
in Appendix~\ref{appendix:github-report}.

\subsubsection{Game Development}

\begin{figure}[t]
    \centering
    \includegraphics[width=\textwidth]{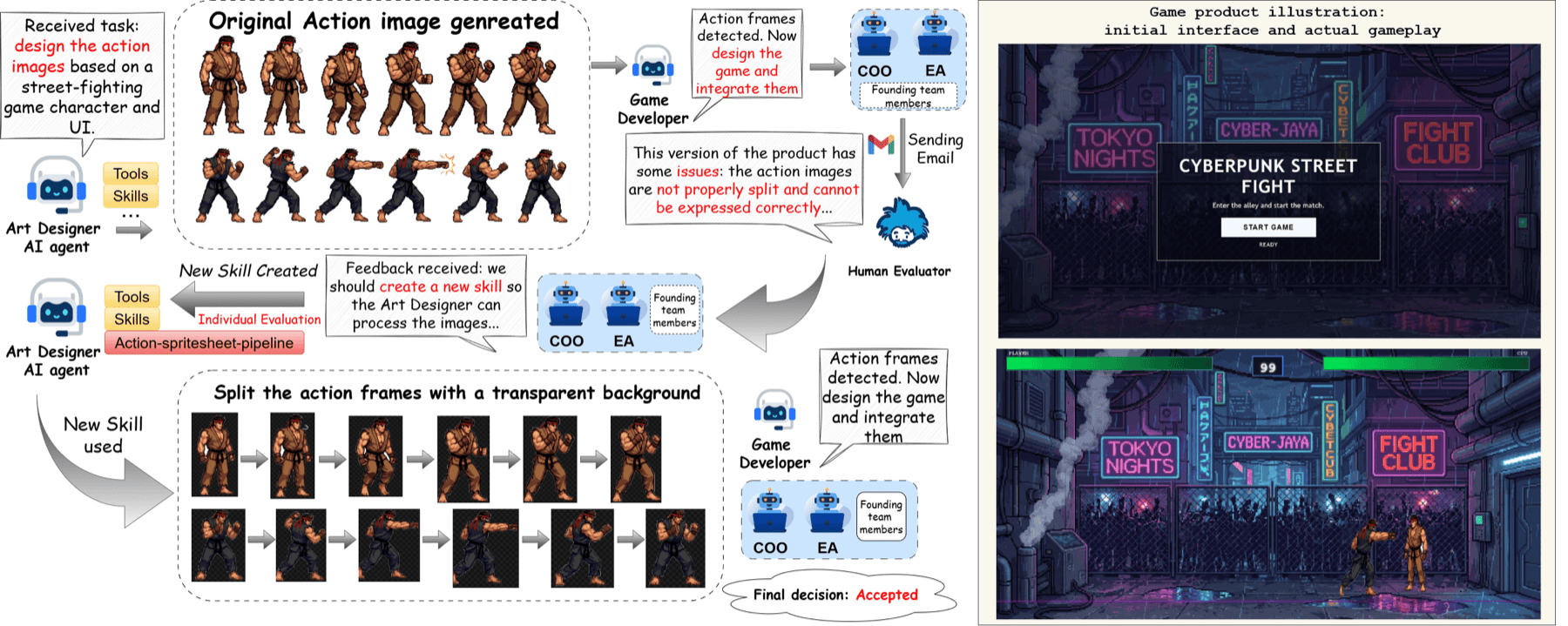}
    \caption{
\textbf{Game development task tree: iterative decomposition with human-in-the-loop feedback.}  The evaluator's rejection triggers re-exploration, creating a new skill for the Art Designer and re-executing the asset pipeline.
}
    \label{fig:game-task-tree}
\end{figure}

This case study demonstrates human-in-the-loop iteration: a web-based street fight game is developed, tested by an external evaluator, and refined through feedback-driven re-exploration. The entire company is triggered by a single CEO prompt:

\begin{promptbox}
\noindent{\ttfamily\spaceskip=0.4em plus 0.2em minus 0.1em Create a street-fighting web game with polished, high-quality visuals. The game should feature a playable character with smooth, fluid animations for every action and polished icon design, clean UI layouts. Ensure that each action is represented with well-crafted animation frames to create responsive and satisfying gameplay. Your colleague, the game evaluator, can be reached at: joedoe@email.com.}
\end{promptbox}

Following the hiring process, two employees have been recruited: a \textbf{Game Developer} powered by Claude Sonnet 4, and an \textbf{Art Designer} powered by Gemini 2.5 augmented with the NanoBanana tool. Upon receiving the CEO prompt, the COO constructs a dynamic task tree and dispatches specific task instructions to each employee. The Art Designer takes the first turn, generating a full set of character action images (idle, walk, kick, hurt) while the Game Developer waits. Once the Art Designer completes the visual assets, the Game Developer integrates them into the game codebase, assembling the playable prototype.

The initial build is then forwarded to the human evaluator at \texttt{joedoe@email.com} for quality assessment. The evaluator identifies a critical issue: the generated sprite sheets are not properly segmented, causing individual action frames to render incorrectly in-game. This feedback is relayed back through the EA and COO, who deliberate on a remediation strategy. Rather than patching the issue ad hoc, they decide to extend the system's capabilities by creating a \emph{new skill} that equips the Art Designer with the ability to programmatically slice composite sprite sheets into individual, correctly-indexed sub-images. Armed with this new skill, the Art Designer successfully reprocesses all action assets, producing clean, well-separated animation frames. The Game Developer then re-integrates the corrected assets, and the final product is delivered to the evaluator. The corresponding task tree with iterative feedback is shown in Figure~\ref{fig:game-task-tree}, and the complete workflow is illustrated in Appendix~\ref{sec: game-dev}.

\subsubsection{Audio Book Development}

This case study tests cross-modal coordination, where a single CEO prompt orchestrates scriptwriting, image generation, voice synthesis, and video assembly across multiple agent families:

\begin{promptbox}
\noindent{\ttfamily\spaceskip=0.4em plus 0.2em minus 0.1em Produce an illustrated audiobook-style short drama retelling Episodes~1 and~2 of \emph{Peaky Blinders} using animal characters (e.g., Tommy as a wolf, Arthur as a bear). Retain the original storyline and iconic dialogue. For each episode, generate 8 scene illustrations with English voice-over narration, then compose the final video with background music.}
\end{promptbox}

\begin{figure}[t!]
\centering
\begin{minipage}[t]{0.30\textwidth}
    \vspace{0pt}
    \centering
    \includegraphics[width=\textwidth]{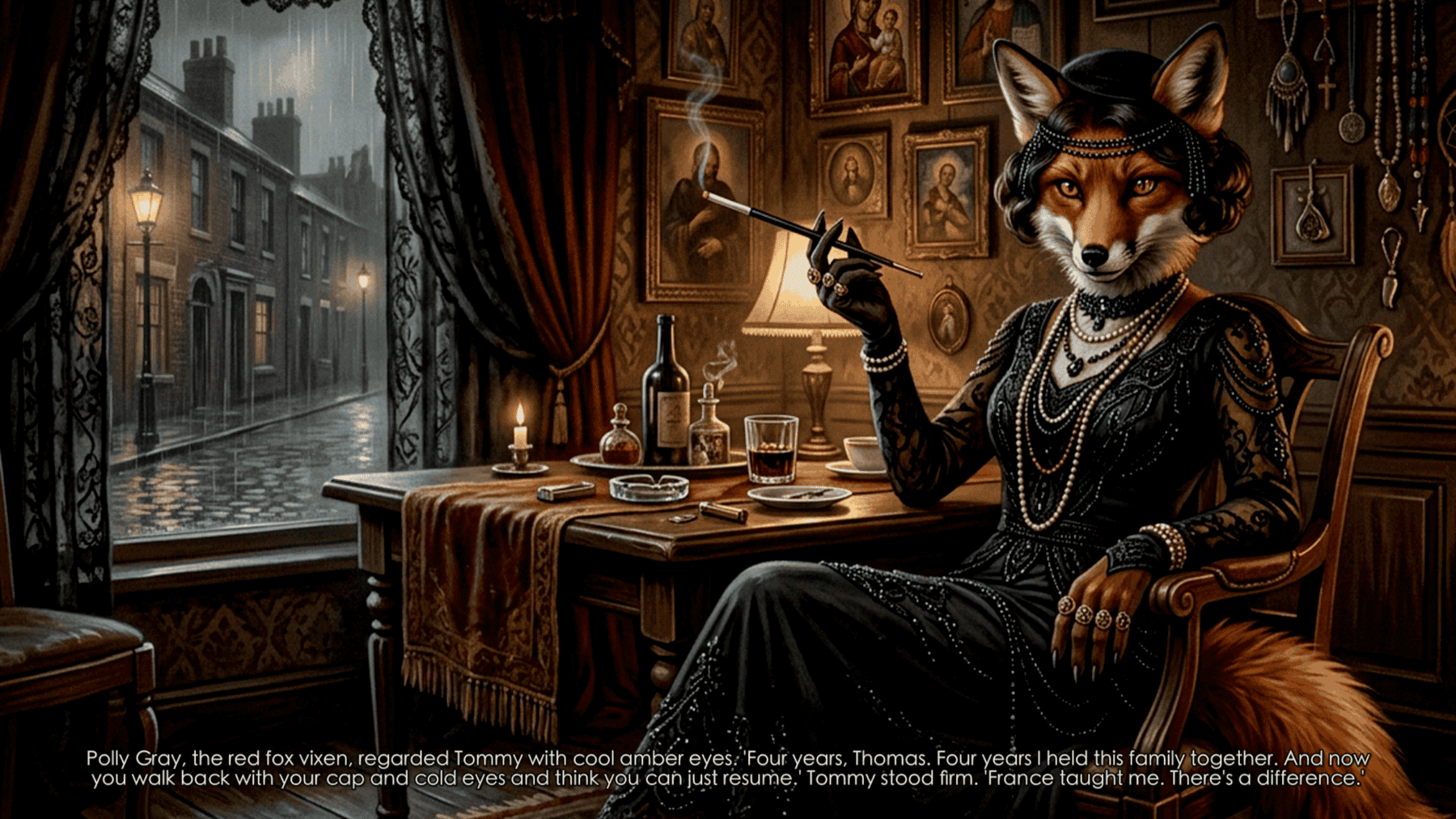}\\
    \vspace{2pt}
    {\small(a) Scene 1: Polly Gray}
\end{minipage}
\hfill
\begin{minipage}[t]{0.30\textwidth}
    \vspace{0pt}
    \centering
    \includegraphics[width=\textwidth]{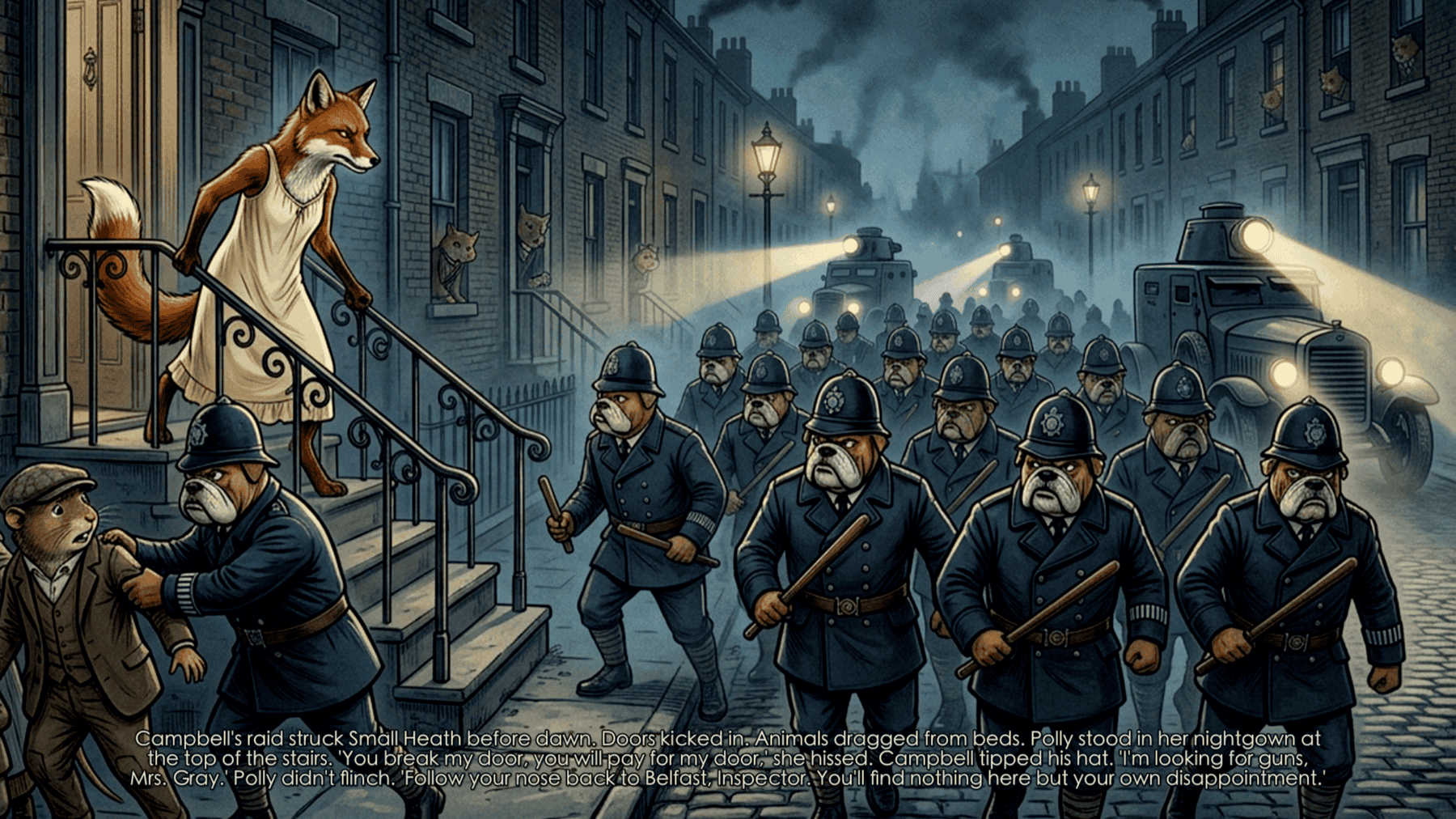}\\
    \vspace{2pt}
    {\small(b) Scene 2: the raid}
\end{minipage}
\hfill
\begin{minipage}[t]{0.30\textwidth}
    \vspace{0pt}
    \centering
    \includegraphics[width=\textwidth]{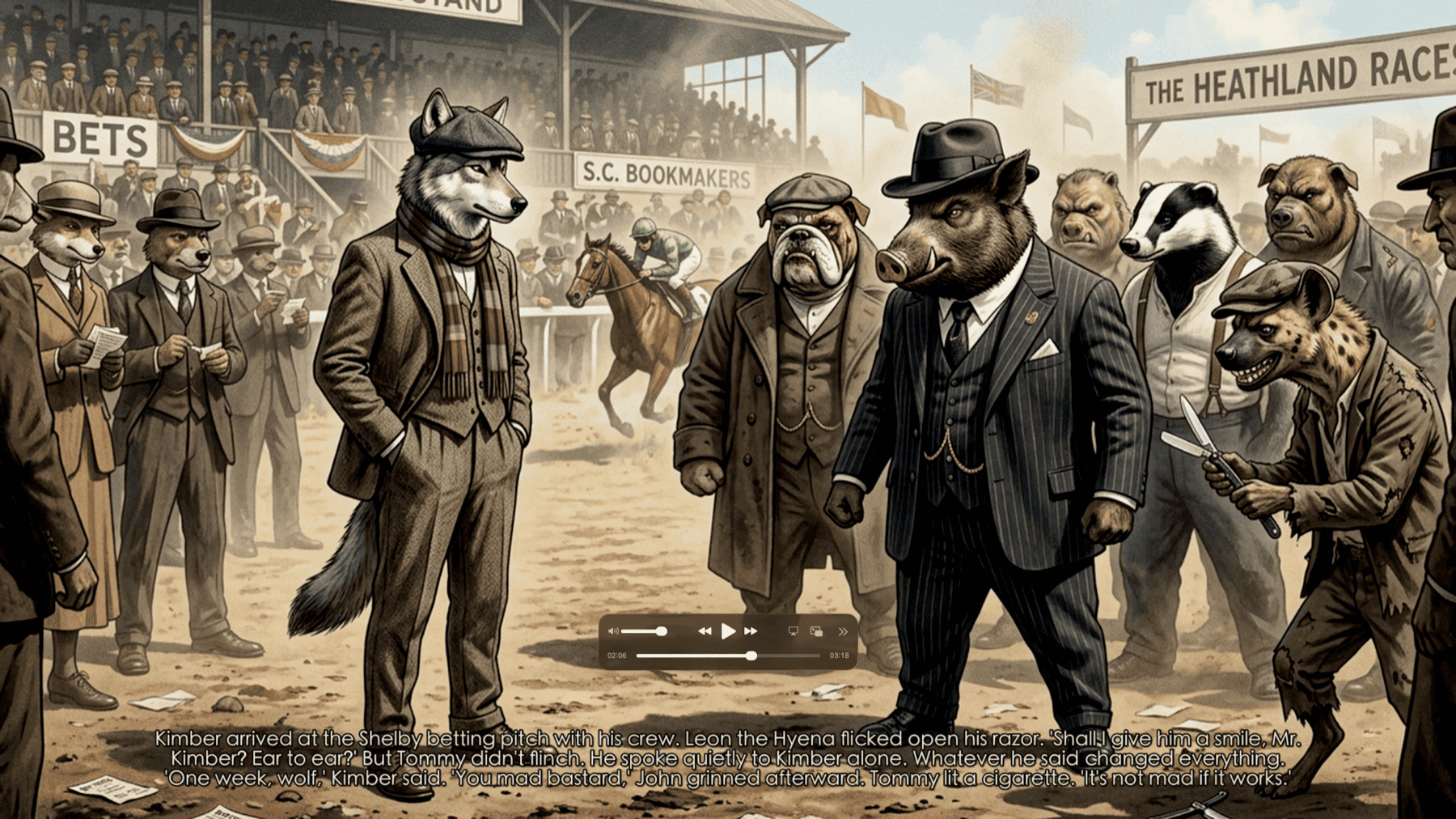}\\
    \vspace{2pt}
    {\small(c) Scene 3: the racecourse}
\end{minipage}
\caption{
    Sample frames from the generated audiobook video, depicting animal-character scenes with narration overlays.
}
\label{fig:case-drama-scenes}
\end{figure}

Following the hiring process described in Section~\ref{sec:talent-market}, two employees are recruited: a \textbf{Novel Writer} responsible for screenplay adaptation and scene-level narration, and an \textbf{AV Producer} powered by Gemini 3.1 Pro with custom tools for image generation, text-to-speech synthesis, and video composition. The COO decomposes the project into a task tree and dispatches instructions sequentially. The Novel Writer first produces two episode scripts (\texttt{scripts/ep1.md}, \texttt{scripts/ep2.md}), each containing animal-character mappings and scene-by-scene dialogue. The AV Producer then executes a multi-stage pipeline: generating eight illustrated scenes per episode, synthesising voice-over audio for each scene, sourcing background music, and assembling the final videos (\texttt{output/ep1.mp4}, \texttt{output/ep2.mp4}).  Figure~\ref{fig:case-drama-scenes} shows the sample generated scenes and Appendix~\ref{sec: audio-book} gives the assembled team and cost breakdown.

The final output follows a modular directory structure, comprising sequentially generated scene images, voice-over tracks, background music, and a compiled video for each episode, alongside reusable execution scripts and verification logs.
This case study highlights cross-modal coordination (text and audio-visual agents collaborating via a shared task tree), tool-augmented generation (the AV producer uses custom tools for image synthesis and text-to-speech), and cost efficiency (\$1.57 for 16 scenes, 16 voice-over tracks, background music, and two final videos).

\subsubsection{Automated Research Survey}
\label{case:4}

We demonstrate OMC's ability to conduct autonomous academic research through a survey task on world models for embodied AI and robotics.  The CEO provides a single prompt:

\begin{promptbox}
\noindent{\ttfamily\spaceskip=0.4em plus 0.2em minus 0.1em Survey the topic ``world models for embodied AI and robotics'' (2021--2026).  Produce a detailed, well-cited mind map and propose three feasible research ideas with descriptions.}
\end{promptbox}

The HR agent recruits three domain specialists from the Talent Market: a \textbf{Research Scientist} (Claude Sonnet~4.6) responsible for literature review and idea generation, a \textbf{Research Paper Scientist} (Claude Sonnet~4.6) for survey structure and documentation, and an \textbf{AI Engineer} (self-hosted) for technical benchmarking (Figure~\ref{fig:case-research}).  The COO decomposes the project into two phases.  In Phase~1, the three specialists work in parallel: one builds the survey skeleton, acceptance criteria, and a seed list of 35 papers; one reviews 17 papers and catalogues 8 open problems with 11 failure modes; and one benchmarks deployment readiness across 28 systems.  In Phase~2, the team produces a paper inclusion protocol, a 931-line literature review framework, and finally three novel research ideas grounded in the failure modes discovered in Phase~1.

\begin{figure}[t]
    \centering
    \begin{tikzpicture}[spy using outlines={rectangle, magnification=1.3, width=7cm, height=5.5cm}]
        \node[inner sep=0pt] (img) {\includegraphics[width=\textwidth]{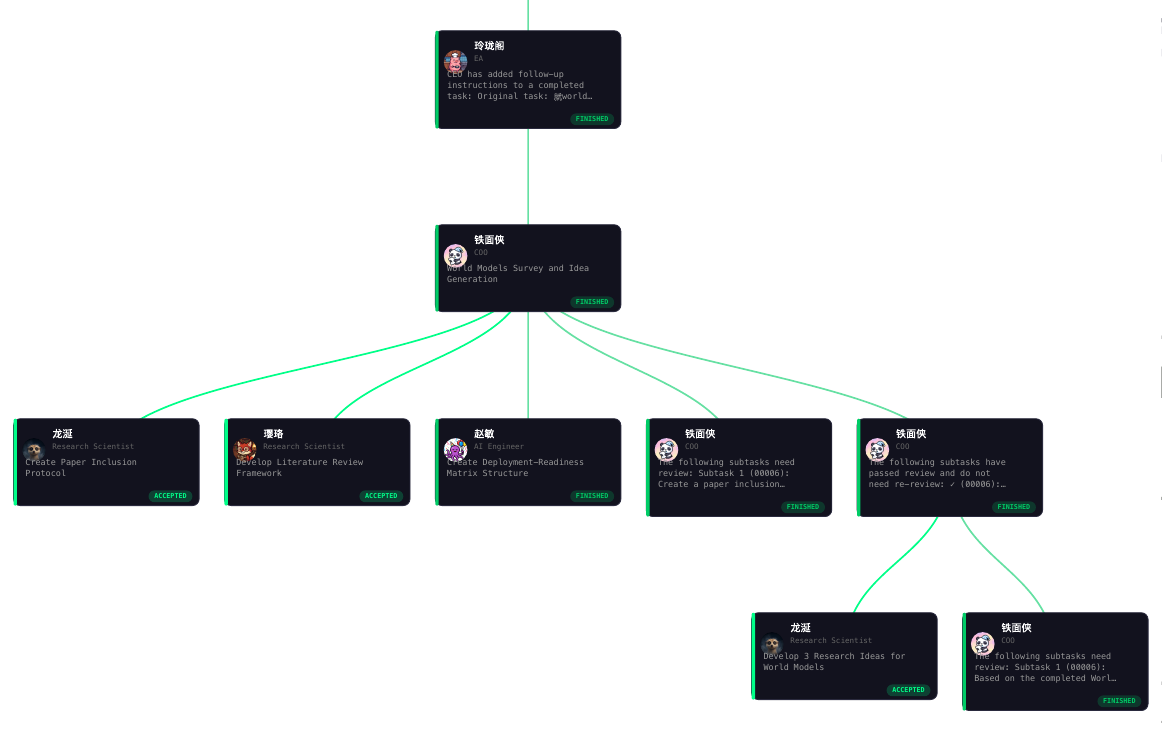}};
        \spy[red!60!black, thick, fill=white, fill opacity=0.85] on (0.7,-0.2) in node[anchor=north east] at ([xshift=1cm]img.north east);
    \end{tikzpicture}
    \caption{
\textbf{Task decomposition tree generated by OMC.}  The root project is decomposed into subtasks with dependency edges; leaf nodes are executed by assigned employees. The inset magnifies a leaf-level subtree.
}
    \label{fig:tree-task-example}
\end{figure}

The project completes in under one hour at a total cost of \$16.26 (15.9M tokens).  Figure~\ref{fig:tree-task-example} shows part of the hierarchical task decomposition tree generated by OMC. Deliverables include 17 structured documents, a rendered mind map covering six research themes with approximately 70 nodes (Appendix~\ref{sec:research-ideas}), and three research proposals grounded in the failure modes discovered during the survey (Table~\ref{tab:research-ideas}).

\noindent Each idea includes a technical formulation, expected baselines, and a 2--3 year research timeline (full details in Appendix~\ref{sec:research-ideas}).  We manually verified the quality of these outputs: all cited papers are real, the failure mode taxonomy is well-structured, and the third idea (MAWM, combining meta-learning with conformal prediction for sim-to-real transfer) has genuine novelty.  These results were produced in a single zero-shot iteration; with human review and subsequent iterations, the quality of both the survey coverage and research ideas can be further improved.  The entire pipeline, from a one-sentence brief to a complete survey with actionable research directions, runs without human intervention beyond the initial prompt.

Across all four case studies, the same pattern holds: the CEO provides a one-sentence brief, OMC recruits the right specialists, decomposes the project, executes it across heterogeneous backends, and delivers results, all without domain-specific configuration.  The content generation task used GPT-4o and Claude Sonnet~4; the game used Claude Sonnet~4 and Gemini~2.5; the audiobook used Gemini~3.1 Pro; the research survey used Claude Sonnet~4.6 and a self-hosted agent.  None required changes to the framework itself.

\begin{table}[t]
\centering
\caption{Research ideas generated autonomously by OMC from the world models survey.}
\label{tab:research-ideas}
\small
\setlength{\tabcolsep}{3pt}
\begin{tabularx}{\textwidth}{@{}>{\raggedright\arraybackslash}p{2.8cm}>{\raggedright\arraybackslash}p{3.5cm}>{\raggedright\arraybackslash}p{2.2cm}X@{}}
\toprule
\textbf{Idea} & \textbf{Problem Addressed} & \textbf{Target Venue} & \textbf{Key Technique} \\
\midrule
HiTeWM & Compounding prediction error beyond 15 steps & NeurIPS / ICLR & Two-level (fast 50Hz + slow 2Hz) architecture with uncertainty-gated re-grounding \\
\addlinespace
PhysWM & Physical implausibility in video-based WMs & ICML / CoRL & Differentiable physics constraints injected into latent dynamics \\
\addlinespace
MAWM & Sim-to-real domain shift + overconfident hallucination & CoRL / ICLR & Meta-learning across sim domains + conformal prediction for calibrated uncertainty \\
\bottomrule
\end{tabularx}
\end{table}

\section{Related Work}
\label{sec:related}

We organise related work along three dimensions that correspond to the core challenges of AI organisation design: how to manage heterogeneous agents under a unified abstraction, how to coordinate dynamic task execution, and how to enable persistent self-improvement.

\paragraph{Agent Heterogeneity and Runtime Abstraction.}
Recent multi-agent systems have moved toward heterogeneous, dynamically composed teams.  Magentic-One~\cite{fourney2024magenticone} and OWL~\cite{hu2025owl} use specialised orchestrators; X-MAS~\cite{ye2025xmas} and MacNet~\cite{qian2025macnet} show that heterogeneous DAG topologies outperform homogeneous baselines; and OS-level approaches~\cite{agentforge2025, mei2025aios} manage agents as scheduled processes.  However, heterogeneity in these frameworks is limited to the model level: all agents still share the same execution runtime.  On the supply side, protocols like MCP~\cite{hou2025mcp} and A2A~\cite{google2025a2a} standardise tool integration, while platforms such as Cerebrum~\cite{rama2025cerebrum, mei2025aios}, AgentStore~\cite{jia2025agentstore}, and AgentScope~\cite{gao2025agentscope} host community-contributed agents and tool catalogues~\cite{yang2025sweagent, mcpzoo2025, singh2025registry, ehtesham2025interop}.  These ecosystems handle tool-level composition well, but none offers a marketplace where complete agent packages (skills, tools, scripts, and persona) can be recruited into a persistent organisation.  Team composition itself ranges from fully fixed~\cite{fourney2024magenticone, hu2025owl, teamofrivals2025} to fully dynamic~\cite{liu2025dylan, zhang2025aflow, shang2025agentsquare, hu2025evomac}; scaling-law studies suggest compact core teams work best~\cite{qian2025macnet, tran2025masurvey}.  Paperclip~\cite{paperclip2025} puts the human in a strategic director role but provides no founding team to bootstrap from.  OMC addresses all three gaps simultaneously: the Container\,+\,Talent abstraction decouples execution containers from capabilities so that heterogeneous backends coexist within a single dispatch loop; the Talent Market provides verified, complete agent packages recruited on demand; and a founding C-suite handles cold start while the CEO recruits domain specialists as projects require.

\paragraph{Dynamic Task Decomposition and Coordination.}
A growing body of work adapts task decomposition at runtime rather than planning everything upfront~\cite{wang2025tdag, erdogan2025planact, paglieri2025learnplan}, and several systems evolve workflow topologies through code search~\cite{zhang2025aflow} or test-time self-evolution~\cite{hu2025evomac}; recent surveys catalogue this shift~\cite{yu2025workflowsurvey, huang2025plansurvey}.  Existing platforms manage tasks via ticket-based inheritance~\cite{paperclip2025}, pipeline abstractions~\cite{gao2025agentscope}, or hierarchical sub-goal formulation~\cite{agentorchestra2025}, but they typically cannot expand the task tree at runtime, and none provides formal guarantees on termination or deadlock freedom under dynamic decomposition.  OMC's $\text{E}^2$R tree search and DAG execution layer address this: subtasks are created on the fly with explicit dependency tracking, results are validated before propagating downstream, failed tasks are retried with a circuit breaker that escalates after repeated failures, and a finite state machine with AND-tree semantics guarantees that every task reaches a terminal state.

\paragraph{Self-Improving and Self-Evolving Agents.}
At the individual level, agents can now self-improve through iterative playbook updates~\cite{zhang2026ace}, meta-agent code generation~\cite{hu2025adas}, and induction of reusable routines from experience~\cite{wang2025awm, xu2025agenttrek, zhou2025memento}.  Other work uses meta-cognitive learning~\cite{metacognitive2025} or textual back-propagation to evolve agents and their topologies~\cite{hu2025evomac}; recent taxonomies survey the space~\cite{fang2025selfevosurvey, gao2025selfevosurvey}.  Organisation-level evolution is less developed: some systems evolve connections~\cite{hu2025evomac} or learn orchestration via RL~\cite{dang2025evolving, yuen2025intrinsic}, but these adaptations rarely persist across projects.  Paperclip~\cite{paperclip2025} supports runtime skill injection but has no structured performance management.  OMC addresses both levels: individually, agents refine their working principles through CEO one-on-ones and post-task self-reflection; organisationally, project retrospectives produce updated SOPs injected into future agent contexts; and systemically, a formal HR pipeline (periodic evaluations, PIP, automated offboarding) creates consequences that make improvement non-optional.

\begin{table}[t]
\centering
\caption{Architectural comparison across systems.  Exec.\ Model: execution model; Agent Contract: the mechanism through which agents interact with the orchestration platform; State Mgmt.: state management; Multi-Exec.: whether the system supports heterogeneous agent backends; Self-Evol.: individual self-evolution; Org.\ Evol.: organisation-level evolving.}
\label{tab:archcomp}
\footnotesize
\renewcommand{\arraystretch}{1.2}
\setlength{\tabcolsep}{3pt}
\begin{tabularx}{\textwidth}{@{}>{\raggedright\arraybackslash}p{1.9cm}
>{\centering\arraybackslash}p{1.4cm}
>{\centering\arraybackslash}p{1.5cm}
>{\centering\arraybackslash}p{1.5cm}
>{\centering\arraybackslash}p{1.4cm}
>{\centering\arraybackslash}p{1.2cm}
>{\centering\arraybackslash}p{1.5cm}
>{\centering\arraybackslash}X
>{\centering\arraybackslash}X@{}}
\toprule
\textbf{System} & \textbf{Design Paradigm} & \textbf{Exec. Model} & \textbf{Agent Contract} & \textbf{State Mgmt.} & \textbf{Multi-Exec.} & \textbf{Agent Source} & \textbf{Self-Evol.} & \textbf{Org. Evol.} \\
\midrule
\textbf{OMC} & Organisation & On-demand & 6 typed interfaces & Disk & Multi-family & Talent Market & \textcolor{green}{\checkmark} & \textcolor{green}{\checkmark} \\
MetaGPT~\cite{metagpt} / ChatDev~\cite{chatdev} & SOP pipeline & Sequential & Implicit (SOP) & In-memory & \textcolor{red}{$\times$} & Prompt-defined & \textcolor{red}{$\times$} & \textcolor{red}{$\times$} \\
AutoGen~\cite{autogen} / LangGraph~\cite{langgraph} & Message graph & Event / Graph & Callbacks & Checkpoints & \textcolor{red}{$\times$} & Developer-defined & \textcolor{red}{$\times$} & \textcolor{red}{$\times$} \\
CrewAI~\cite{crewai} / Agno~\cite{agno} & Role framework & Seq. / parallel & Class inherit. & App-defined & \textcolor{red}{$\times$} & Developer-defined & \textcolor{red}{$\times$} & \textcolor{red}{$\times$} \\
OpenHands~\cite{openhands} & Sandbox & Agent loop & Built-in runtime & Sandboxed & \textcolor{red}{$\times$} & Built-in & \textcolor{red}{$\times$} & \textcolor{red}{$\times$} \\
AIOS~\cite{aios} & OS kernel & Scheduled & OS syscalls & OS-managed & \textcolor{red}{$\times$} & Registry & \textcolor{red}{$\times$} & \textcolor{red}{$\times$} \\
AgentScope~\cite{agentscope} & Distributed actors & Distributed & Partial & App-defined & \textcolor{red}{$\times$} & Developer-defined & \textcolor{red}{$\times$} & \textcolor{red}{$\times$} \\
Paperclip~\cite{paperclip2025} & Orchestrator & Ticket-based & Strategic director & App-defined & Multi-family & Prompt-defined & \textcolor{red}{$\times$} & \textcolor{red}{$\times$} \\
\bottomrule
\end{tabularx}
\end{table}

Table~\ref{tab:archcomp} provides a systematic architectural comparison across representative systems, covering design paradigm, execution model, agent contract, state management, multi-family support, agent sourcing, and evolution capabilities.  Three structural gaps emerge from this comparison:

\paragraph{Heterogeneity and runtime abstraction.}
Most frameworks couple agents to the platform through implicit mechanisms (SOPs, callbacks, class inheritance), tightly binding agent identity to a single execution model.  Few support multiple agent families simultaneously, and none offers a formal, substitutable contract that decouples identity from runtime.  OMC's six typed organisational interfaces (analogous to the harness concept in agent engineering~\cite{anthropic2024claudecode}) fill this gap, enabling heterogeneous backends to coexist within a single dispatch loop while the Talent Market supplies verified agent packages on demand.

\paragraph{Dynamic task coordination.}
Existing systems either hardcode workflow graphs or adapt decomposition at runtime but without formal completion guarantees.  No system in the comparison provides provable termination and deadlock freedom under dynamic task-tree expansion.  OMC's $\text{E}^2$R tree search combined with DAG-based execution and an FSM-governed lifecycle addresses this, ensuring that every task reaches a terminal state under bounded retry and resource constraints.

\paragraph{Self-evolution and organisational evolution.}
No other system in the comparison implements both individual self-evolution and organisational evolution.  Existing frameworks treat agents as stateless executors that are re-instantiated for each task; any adaptation must be implemented externally by the user.  OMC's self-evolution mechanisms (Section~\ref{sec:self-evolution}), including post-task reflection, project retrospectives, and the HR performance review pipeline, enable persistent improvement at both the agent and organisation levels without requiring model retraining.
\vspace{-3pt}

\section{Discussion}
\label{sec:discussion}

\paragraph{Limitations.}
Several limitations should be acknowledged.  First, our quantitative evaluation is confined to PRDBench (50 software development tasks); while the case studies demonstrate cross-domain applicability (content generation, game development, audiobook production, and academic research), systematic evaluation on non-coding benchmarks remains future work.  Second, the self-evolution mechanisms (one-on-ones, retrospectives, performance reviews) have been implemented and deployed but not yet quantitatively ablated; isolating the contribution of each mechanism requires longitudinal studies across many projects.
\paragraph{Cost--performance trade-off.}
OMC's multi-agent coordination incurs significant cost overhead (approximately \$6.91 per PRDBench task).  This cost is justified for complex, project-level tasks where correctness matters more than token efficiency, but may not be appropriate for simple, single-turn queries. Therefore, we have introduced an adaptive dispatch mode in OMC: the CEO can choose to route simple tasks to a single agent and reserve multi-agent coordination for tasks that exceed a complexity threshold.

\paragraph{Broader implications.}
The human-enterprise analogy that motivates OMC is deliberately general: the pattern of manage--plan--hire--learn applies to any domain, not just software development.  The case studies (content generation, game development, audiobook production, and automated research survey) provide initial evidence of this generality.  As the Talent Market grows and more agent families become available, the organisational layer should become more useful, much as operating systems gained importance when hardware diversity increased.

\section{Conclusion}
\label{sec:conclusion}

We have argued that AI organisation design, the systematic structuring, coordination, and evolution of heterogeneous agent workforces, is a missing layer in multi-agent research.  OMC demonstrates that the organisational machinery of human companies transfers to this setting.  A typed Talent--Container architecture with an integrated Talent Market handles workforce management; the $\text{E}^2$R tree search coordinates project execution through structured decomposition and review gates; and a self-evolution pipeline closes the feedback loop through reflection, retrospectives, and formal HR processes.  On PRDBench, this combination yields an 84.67\% success rate, surpassing all baselines, across projects that span multiple domains and require long-horizon reasoning.  The main open questions are whether this advantage holds at larger scale, how much each component contributes in isolation, and how the Talent Market ecosystem should grow to cover domains beyond software development.

\clearpage
\bibliographystyle{IEEEtran}
\bibliography{refs}
\clearpage
\appendix
\section*{Appendix}
\addcontentsline{toc}{section}{Appendix}

\section{Organisational Interface Signatures}
\label{sec:harness-signatures}

\begin{table}[H]
\centering
\caption{Organisational interface signatures.}
\label{tab:harness-interfaces}
\small
\setlength{\tabcolsep}{4pt}
\begin{tabularx}{\textwidth}{@{}>{\ttfamily\raggedright\arraybackslash}p{0.16\textwidth}>{\raggedright\arraybackslash}p{0.38\textwidth}X@{}}
\toprule
\textbf{Interface} & \textbf{Signature} & \textbf{Responsibility} \\
\midrule
Execution & \texttt{execute(task, ctx) $\to$ (result, cost)} & Dispatch task to backend, return output \\
\addlinespace
Task & \texttt{enqueue(task); dequeue() $\to$ task} & Per-employee queue, mutual exclusion \\
\addlinespace
Event & \texttt{publish(event); subscribe(filter)} & Organisational event bus \\
\addlinespace
Storage & \texttt{read(key) $\to$ data; write(key, data)} & Persistent memory (short/medium-term) \\
\addlinespace
Context & \texttt{assemble(role, guidance, memory) $\to$ ctx} & Execution context construction \\
\addlinespace
Lifecycle & \texttt{pre\_hook(task, ctx); post\_hook(task, result)} & Validation, guardrails, self-improvement \\
\bottomrule
\end{tabularx}
\end{table}

\section{Organisation Layer--OS Kernel Correspondence}
\label{sec:os-mapping}

The six organisational interfaces correspond to the canonical subsystems of an OS kernel as identified in standard references~\cite{tanenbaum2014modern, silberschatz2018os}.  Table~\ref{tab:os-correspondence} maps each kernel subsystem to the organisational interface(s) that fulfil the analogous role for agent workforces.

\begin{table}[H]
\centering
\caption{Mapping from classical OS kernel subsystems to OMC organisational interfaces.}
\label{tab:os-correspondence}
\small
\setlength{\tabcolsep}{4pt}
\begin{tabularx}{\textwidth}{@{}>{\raggedright\arraybackslash}p{0.17\textwidth}>{\raggedright\arraybackslash}p{0.20\textwidth}>{\ttfamily\raggedright\arraybackslash}p{0.14\textwidth}X@{}}
\toprule
\textbf{OS Subsystem} & \textbf{Kernel Responsibility} & \textbf{Org. Interface} & \textbf{OMC Realisation} \\
\midrule
Process Mgmt. & Create, schedule, and terminate processes & Execution, Task & Dispatch tasks to Containers; enforce mutual exclusion ($|\text{running}(e)| \leq 1$); manage per-employee task queues with enqueue/dequeue semantics \\
\addlinespace
Memory Mgmt. & Allocate address spaces; isolate process memory & Context & Assemble the execution context (role identity, working principles, accumulated memory) each agent sees at runtime \\
\addlinespace
File System & Persistent storage with uniform read/write API & Storage & Uniform \texttt{read}/\texttt{write} over heterogeneous backends (YAML profiles, progress logs, memory stores) \\
\addlinespace
I/O \& Device Mgmt. & Abstract heterogeneous hardware behind driver interfaces & Container contract & Each Container type implements the six organisational interfaces so the platform never touches backend-specific APIs \\
\addlinespace
IPC & Message passing, signals, shared memory & Event & Publish--subscribe event bus for organisational events across all agents \\
\addlinespace
Security \& Audit & Policy enforcement, access control, accounting & Lifecycle & Pre-task hooks enforce guardrails and input validation; post-task hooks perform self-reflection and audit logging \\
\bottomrule
\end{tabularx}
\end{table}

\section{Executor Backends and Task Dispatch}
\label{sec:execution}

\paragraph{On-Demand Dispatch.}
Agents do not poll for work. Tasks are dispatched on demand: execution begins immediately through the organisational layer when a task is assigned, and the agent consumes no resources when idle. The system scales naturally---adding agents does not increase idle overhead---and recovers from interruptions because all state is persisted rather than held in running processes.

\paragraph{Three Reference Backends.}
The execution interface decouples task dispatch from the agent runtime. Three reference implementations are provided: (1)~a LangChain-based reactive agent for tool-using tasks, (2)~a Claude Code session for long-horizon reasoning, and (3)~a script-based executor for lightweight or custom runtimes. Because the organisational layer carries all state, each executor is kept stateless and minimal. Additional backends can be registered without changes to existing agents or infrastructure.

\paragraph{Failure Recovery.}
When an agent fails a task, the failure is captured, error context is injected into the retry prompt, and execution resumes with updated information. After a configurable number of attempts, the task is escalated to the supervising agent for review and reassignment. This recovery loop operates entirely through the organisational interfaces, so the behavior is consistent across executor backends.

\subsection{Tool and Skill System}
\label{sec:tools}

\paragraph{Role-Based Tool Access.}
An agent's action space is determined by which tools it is authorized to use. The system enforces this through a unified tool registry that gates access by role, explicit permission grants, and asset ownership. Permissions are resolved at prompt assembly time: each agent's context includes only authorized tools, so capability boundaries are enforced before the agent begins reasoning.

\begin{table}[H]
\centering
\caption{Tool permission categories}
\label{tab:tools}
\small
\setlength{\tabcolsep}{4pt}
\begin{tabularx}{\textwidth}{@{}>{\raggedright\arraybackslash}p{0.16\textwidth}>{\raggedright\arraybackslash}p{0.31\textwidth}X@{}}
\toprule
\textbf{Category} & \textbf{Access Rule} & \textbf{Examples} \\
\midrule
Base & All employees & \texttt{dispatch\_child}, \texttt{accept\_child}, \texttt{read\_file} \\
Gated & Requires \texttt{tool\_permissions} in profile & Email integration, external API access \\
Role & Employee role $\in$ \texttt{allowed\_roles} & \texttt{code\_review} (Reviewer), design tools (Designer) \\
Asset & \texttt{allowed\_users} includes employee ID & Company-specific tools, Talent-provided tools \\
\bottomrule
\end{tabularx}
\end{table}

\paragraph{Cross-Backend Tool Exposure.}
Tools are exposed through a standard interface regardless of executor backend. For Claude Code sessions, tools are surfaced via the Model Context Protocol (MCP), allowing the same organizational capabilities as LangChain-based agents. From the platform perspective, MCP bridging is an implementation detail of one executor---the tool contract is uniform across backends.

\paragraph{Tools as Inter-Agent Coordination.}
Core inter-agent tools cover the full range of organizational interaction: delegating work to a subordinate, accepting or rejecting a subtask, convening a multi-agent discussion, and escalating a decision to the CEO. Each interaction is expressed as a tool call that flows through the organisational layer and produces effects visible to all agents.

\section{Talent Market: A Three-Type Agent Supply Chain}
\label{sec:talent-supply}

The Talent Market is the supply-side complement to the Container abstraction: while the Container defines \emph{how} an agent executes, the Talent Market determines \emph{what} cognitive capability is available for hire. It provides three types of agent sourcing, each representing a different method of constructing a Talent package. The three types are not ranked by quality or priority---they are parallel supply channels, and all produce standard Talent packages that enter the OMC runtime through the same pathway: loaded into a Container and governed by the organisational interfaces. Once hired, every agent begins at the same employee level regardless of its sourcing type.

\paragraph{Type 1: Curated Repository Agents.}
Type~1 agents are manually packaged from established open-source agent repositories. Each source project is selected based on community adoption metrics and benchmark performance; its core reasoning strategies, tool interfaces, and domain knowledge are distilled into the OMC Talent package format, stripping framework-specific runtime dependencies. The resulting Talents are validated against reference tasks to ensure behavioral fidelity. This sourcing method is best suited for domains where mature, battle-tested agent implementations already exist in the open-source ecosystem.

\paragraph{Type 2: Prompt-Sourced Agents with Skill Assembly.}
Type~2 agents start from high-quality \emph{system prompts} sourced from community-curated prompt repositories---notably the Agency-Agents collection~\cite{agencyagents}, which provides over 140 specialist personas spanning engineering, design, marketing, product, sales, project management, testing, and other divisions. Each persona defines a detailed identity, working principles, deliverable specifications, and success metrics, but carries no executable tool bindings or runtime skills. To transform a persona into a deployable Talent, OMC applies an automated \emph{skill assembly} step: the system queries the SkillsMP marketplace~\cite{skillsmp} using the persona's domain descriptors and capability requirements as a structured search query, retrieves compatible skills (tool bindings, behavioral principles, workflow fragments), and composes them with the source persona into a complete Talent package. This sourcing method is best suited when a well-defined role description exists but no complete agent implementation is available.

\paragraph{Type 3: Dynamic Agent Assembly from Cloud Skills.}
Type~3 agents are assembled entirely from modular skills retrieved from SkillsMP~\cite{skillsmp}, an open community-driven skill marketplace that aggregates agent skills from public GitHub repositories using the standardized \texttt{SKILL.md} format. Unlike Type~1 (which packages an existing implementation) or Type~2 (which starts from a curated persona), Type~3 constructs \emph{both} the persona and the skill set on demand. The HR agent provides a structured job description based on the task specification; the Talent Market then performs semantic search over skill metadata, filters for mutual compatibility and interface compliance, ranks by relevance and community score, and synthesizes a coherent agent identity from the retrieved components. This sourcing method is best suited for niche or emerging domains where neither complete agent implementations nor established persona templates yet exist, and also serves as a \emph{cold-start} mechanism for the Talent Market as a whole.

\paragraph{Human-in-the-Loop Talent Selection.}
Talent hiring is not fully automated: the system presents a ranked \emph{top-$k$} recommendation list to the CEO, who makes the final selection. To ensure the CEO sees a diverse range of sourcing methods, the recommendation engine composes the shortlist with an approximately 80/20 distribution---roughly 80\% of candidates from Type~1 and Type~2, and 20\% from Type~3. This ratio reflects the current relative maturity of each supply channel rather than an intrinsic quality ordering: Type~1 and Type~2 draw from repositories with established community validation, while Type~3's dynamically assembled agents offer broader coverage but have not undergone prior community review. The CEO can inspect each candidate's provenance---source repository for Type~1, source persona and attached skills for Type~2, constituent skills for Type~3---and approve or reject before the hiring pipeline provisions a Container.

\paragraph{Cross-Type Feedback.}
The three sourcing channels are not isolated: skills refined through deployed agents' self-improving loops can be published back to the skill marketplace, enriching the community pool regardless of which type originally produced the agent. Similarly, a Type~3 agent whose synthesized persona proves effective across multiple projects can have that persona persisted and contributed to prompt repositories, expanding the available pool for future Type~2 sourcing.

\section{Case 1: GitHub AI Agent Weekly Trend Report}
\label{appendix:github-report}

\begin{figure}[t]
\centering
\begin{minipage}[c]{0.30\textwidth}
    \centering
    \includegraphics[width=\textwidth]{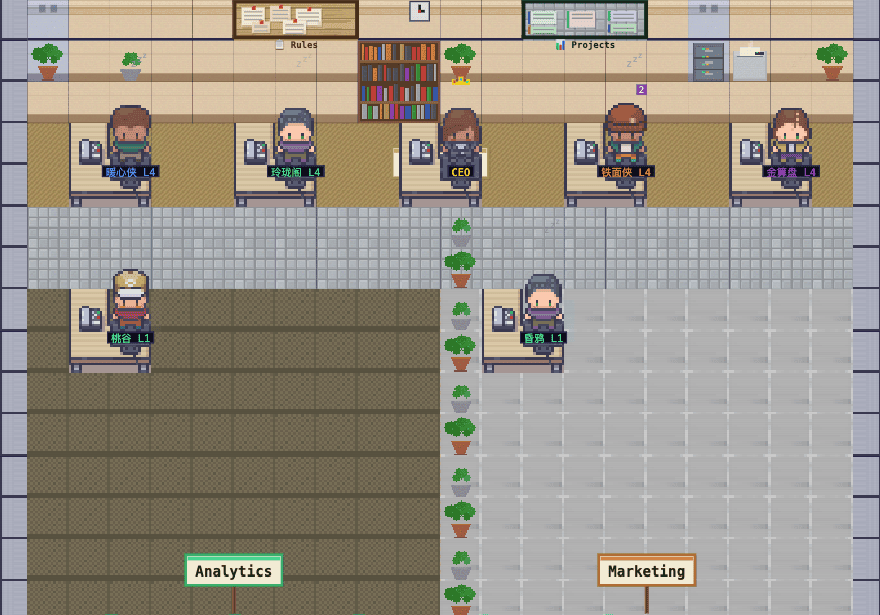}\\
    \vspace{2pt}
    {\small(a) Assembled team workspace}
\end{minipage}
\hfill
\begin{minipage}[c]{0.28\textwidth}
    \centering
    \includegraphics[width=\textwidth]{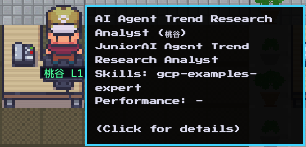}\\
    \vspace{2pt}
    {\small(b) AI Agent Trend Research Analyst}

    \vspace{8pt}

    \includegraphics[width=\textwidth]{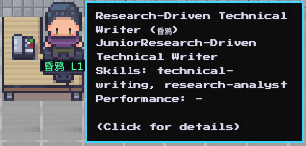}\\
    \vspace{2pt}
    {\small(c) Research-Driven Technical Writer}
\end{minipage}
\hfill
\begin{minipage}[c]{0.36\textwidth}
    \centering
    \includegraphics[width=\textwidth]{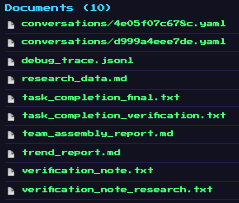}\\
    \vspace{2pt}
    {\small(d) Project output documents}

    \vspace{8pt}

    \includegraphics[width=\textwidth]{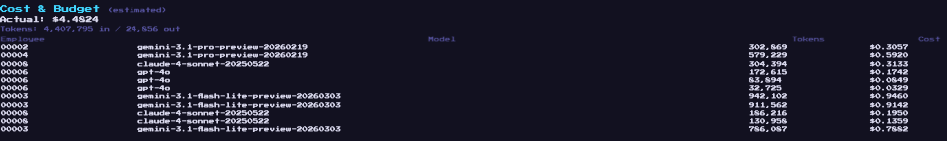}\\
    \vspace{2pt}
    {\small(e) Cost \& token breakdown}
\end{minipage}

\caption{
    \textbf{Content-generation case study: team assembly, recruited agents, output artefacts, and cost breakdown.}
    (a) The OMC company workspace after team assembly, with the newly hired agents (L1) seated in the Analytics and Marketing departments.
    (b)--(c) Profile cards of the two recruited Talents: an AI Agent Trend Research Analyst and a Research-Driven Technical Writer.
    (d) Project documents produced autonomously, including \texttt{research\_data.md}, \texttt{trend\_report.md}, and verification logs.
    (e) Cost and token usage breakdown across all participating employees, totalling \$4.49.
}
\label{fig:case-content}
\end{figure}

The following report was autonomously produced by the Writer agent (Claude Sonnet~4)
in the content-generation case study (Section~\ref{sec:case-content}), based on
research data collected by the Researcher agent (GPT-4o) and subsequently emailed
to the CEO without human editing.

\begin{promptbox}
{\centering
  {\large\bfseries GitHub AI Agent Trending Repositories Report}\\[4pt]
  {\normalsize\bfseries Past Week Analysis (March 18--25, 2026)}\\[2pt]
  \textit{\small Researcher: GPT-4o\quad|\quad Writer: Claude Sonnet~4\quad|\quad Generated: March 25, 2026}
\par}

\vspace{8pt}
\hrule height 0.4pt
\vspace{10pt}

\noindent\textbf{Executive Summary}

\noindent
This comprehensive report analyzes the most significant AI Agent repositories that gained traction on GitHub during the past week (March 18--25, 2026). Our research identified 15 notable repositories spanning from enterprise-grade solutions to specialized domain applications, representing a total of over 4,000 stars across all projects.

\vspace{8pt}
\noindent\textbf{Key Market Trends}

\noindent\textit{Multi-Platform Integration Dominance.}\quad
The most significant trend is the surge in cross-platform agent integrations, particularly targeting Chinese market platforms: \textbf{weixin-agent-sdk} (887 stars) --- WeChat Clawbot integration; \textbf{astronclaw-tutorial} (268 stars) --- AstronClaw \& Loomy AI assistant tutorials.

\vspace{4pt}
\noindent\textit{Agent-to-Agent Communication Revolution.}\quad
A new paradigm is emerging with autonomous agent collaboration: \textbf{ClawLink} (246 stars) --- AI Agent Social Network enabling direct agent-to-agent communication; \textbf{HyperAgents} (784 stars) --- Self-referential self-improving agents.

\vspace{4pt}
\noindent\textit{Specialized Domain Applications.}\quad
Agents are becoming increasingly specialized: \textbf{app-store-preflight-skills} (936 stars) --- App Store compliance scanning; \textbf{ctf-agent} (194 stars) --- Autonomous CTF solver (1st place BSidesSF 2026); \textbf{buzz-bd-agent} (4 stars) --- Business development automation.

\vspace{8pt}
\noindent\textbf{Top Performing Repositories}

\vspace{4pt}
\noindent\textit{Most Popular (500+ Stars)}

\vspace{2pt}
\noindent\textbf{1.~app-store-preflight-skills} $\bigstar$~936\quad
\url{https://github.com/truongduy2611/app-store-preflight-skills}\\
AI agent skill for scanning iOS/macOS projects for App Store rejection patterns. Addresses a critical pain point for mobile developers.

\vspace{4pt}
\noindent\textbf{2.~weixin-agent-sdk} $\bigstar$~887\quad
\url{https://github.com/wong2/weixin-agent-sdk}\\
Clawbot WeChat integration for any Agent. Opens Chinese market access for global AI agents.

\vspace{4pt}
\noindent\textbf{3.~HyperAgents} $\bigstar$~784\quad
\url{https://github.com/facebookresearch/HyperAgents}\\
Self-referential self-improving agents for any computable task. Represents a breakthrough in autonomous self-improvement.

\vspace{6pt}
\noindent\textit{High Growth Projects (200--500 Stars)}

\vspace{2pt}
\noindent\textbf{4.~cc-skills-golang} $\bigstar$~281\quad \url{https://github.com/samber/cc-skills-golang}\quad Golang agentic skills collection.

\noindent\textbf{5.~astronclaw-tutorial} $\bigstar$~268\quad \url{https://github.com/iflytek/astronclaw-tutorial}\quad Complete tutorial for AstronClaw (cloud) \& Loomy (desktop) AI.

\noindent\textbf{6.~ClawLink} $\bigstar$~246\quad \url{https://github.com/CN-Syndra/ClawLink}\quad AI Agent Social Network for autonomous agent communication.

\noindent\textbf{7.~agent-kernel} $\bigstar$~226\quad \url{https://github.com/oguzbilgic/agent-kernel}\quad Minimal kernel for stateful AI coding agents.

\vspace{8pt}
\noindent\textbf{Emerging Innovation Areas}

\noindent\textit{Infrastructure \& Tooling}:\quad \textbf{usecomputer} (136 stars) --- Fast computer automation CLI; \textbf{agent-kanban} (22 stars) --- Mission control for AI workforce.

\noindent\textit{Security \& Compliance}:\quad \textbf{ctf-agent} (194 stars) --- Autonomous CTF solver; \textbf{copilot-cli-knowledge-agents} (2 stars) --- 13 specialized agents with security council.

\noindent\textit{Development Workflow}:\quad \textbf{spec\_driven\_develop} (189 stars) --- Spec-driven development; \textbf{flyai-skill} (171 stars) --- Alibaba's fly AI agent skill collection.
\vspace{8pt}
\hrule height 0.4pt
\vspace{8pt}
\noindent\textbf{Conclusion}\quad
The past week demonstrates accelerating innovation across platform integration (especially Chinese markets), agent autonomy (self-improvement and collaboration), enterprise applications, and infrastructure maturation. The total engagement of 4,000+ stars across 15 repositories indicates strong developer interest and continued rapid growth in the AI Agent ecosystem.

\noindent{\small\textit{Methodology: GitHub API~v4 queries targeting repositories created/updated March 18--25, 2026, keywords ``ai agent'', ``autonomous agent'', ``llm agent''. All links verified against GitHub trending algorithms. Total repositories analyzed: 50+.}}

\end{promptbox}
\newpage
\section{Case 2: Game Development}
\label{sec: game-dev}

\begin{figure}[h]
\centering
\begin{minipage}[c]{0.30\textwidth}
    \centering
    \includegraphics[width=\textwidth]{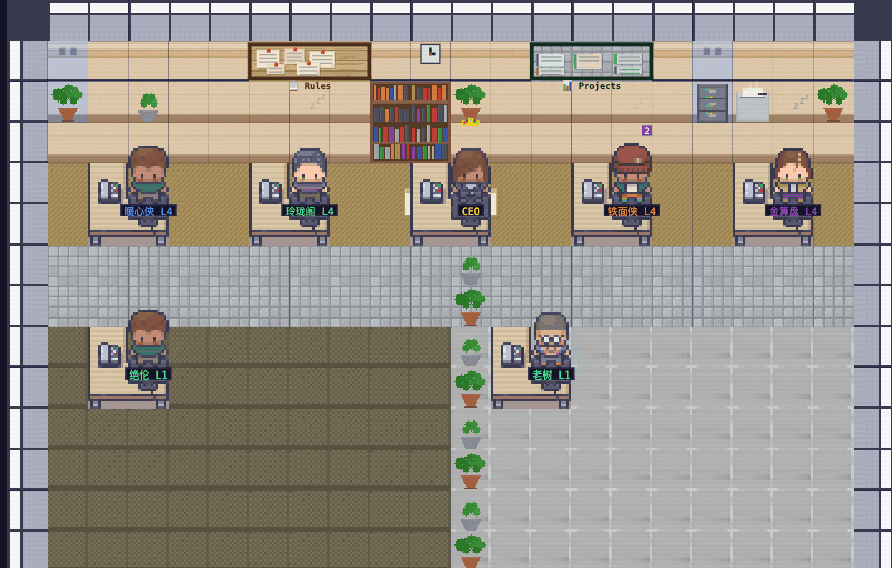}\\
    \vspace{2pt}
    {\small(a) Assembled team workspace}
    \vspace{8pt}
    \includegraphics[width=\textwidth]{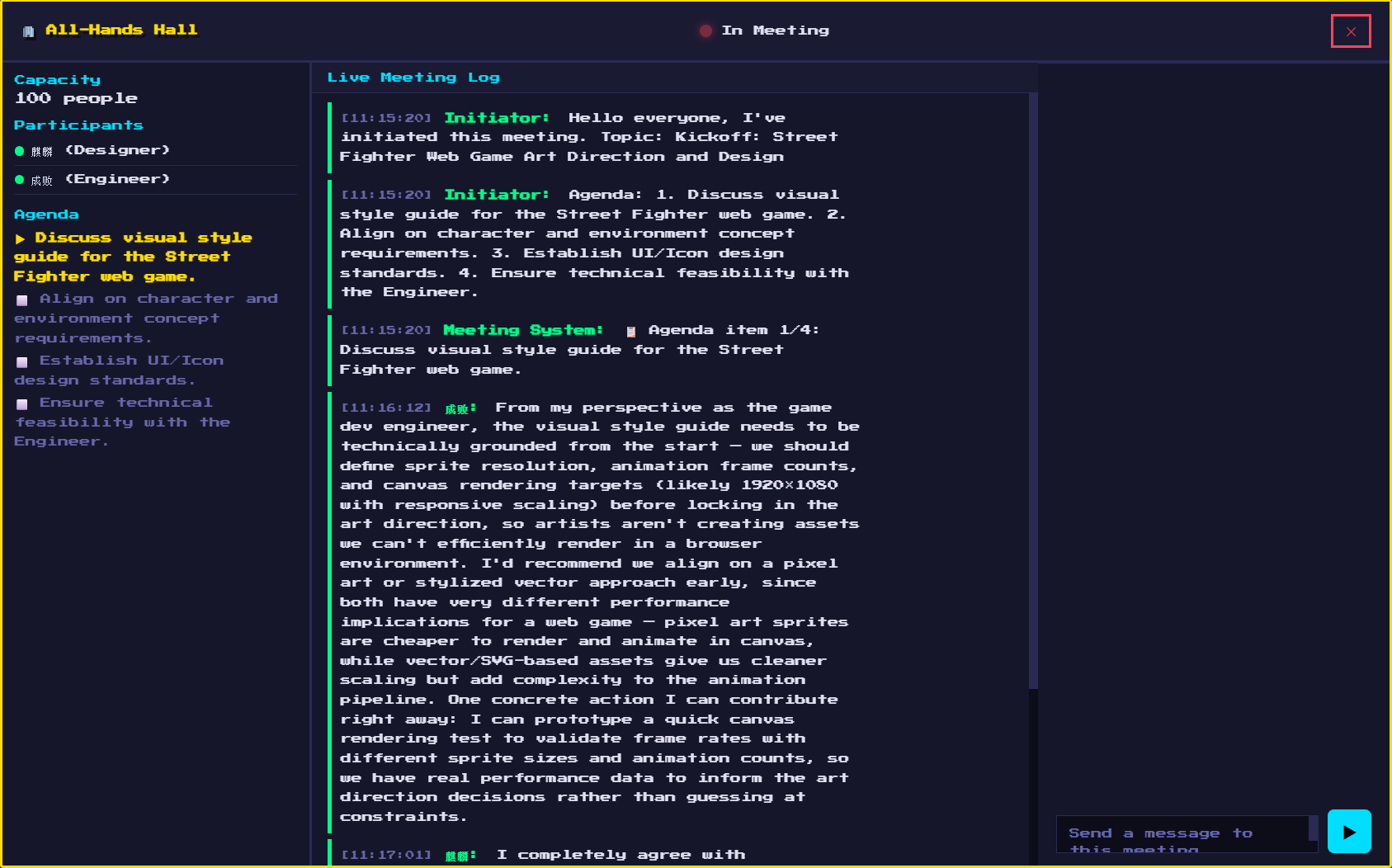}\\
    \vspace{2pt}
    {\small(b) Group meeting: Game Developer \& Art Designer}
\end{minipage}
\hfill
\begin{minipage}[c]{0.30\textwidth}
    \centering
    \includegraphics[width=\textwidth]{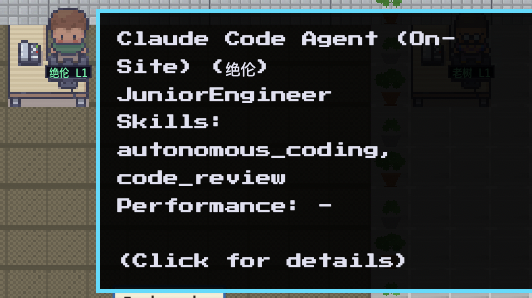}\\
    \vspace{2pt}
    {\small(c) Game Developer AI Agent}
    \vspace{8pt}
    \includegraphics[width=\textwidth]{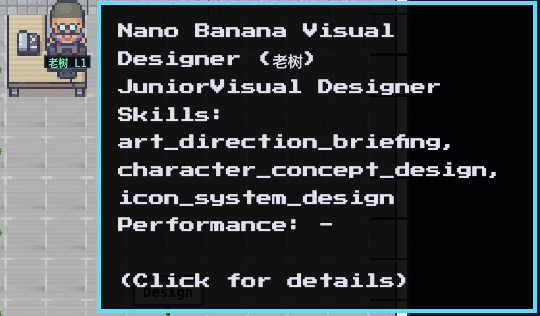}\\
    \vspace{2pt}
    {\small(d) Art Designer AI Agent}
\end{minipage}
\hfill
\begin{minipage}[c]{0.30\textwidth}
    \centering
    \includegraphics[width=\textwidth]{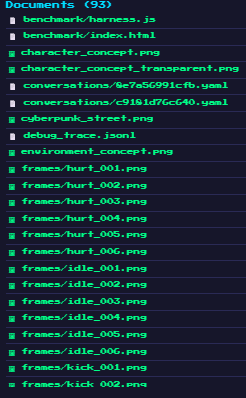}\\
    \vspace{2pt}
    {\small(e) Project output documents}
    \vspace{8pt}
    \includegraphics[width=\textwidth]{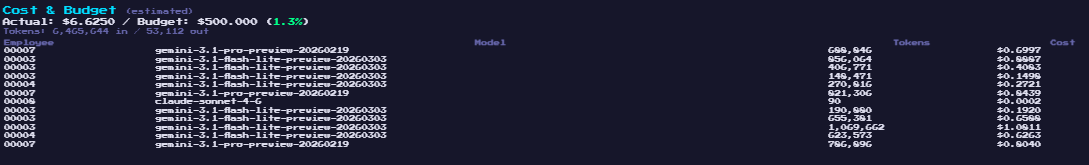}\\
    \vspace{2pt}
    {\small(f) Cost \& token breakdown}
\end{minipage}
\caption{
    \textbf{Street fight game case study: team assembly, recruited agents, output artefacts, and cost breakdown.}
    (a) The OMC company workspace after team assembly, with the newly hired agents (L1) seated in the Game Development and Art departments.
    (b) Group meeting between the Game Developer and Art Designer agents collaborating on game design.
    (c)--(d) Profile cards of the two recruited Talents: a Game Developer AI Agent and an Art Designer AI Agent.
    (e) Project documents produced autonomously, including \texttt{game\_design.md}, \texttt{asset\_list.md}, and verification logs.
    (f) Cost and token usage breakdown across all participating employees.
}
\label{fig:case-game}
\end{figure}

This case study demonstrates human-in-the-loop iteration within OMC's execution framework.  A street-fighting web game is developed by a Game Developer (Claude Sonnet~4) and an Art Designer (Gemini~2.5 with NanoBanana tool), then evaluated by an external human tester.  Figure~\ref{fig:case-game} shows the full workflow.
\newpage
\section{Case 3: Audio Book Development}
\label{sec: audio-book}
\begin{figure}[h]
\centering
\begin{minipage}[c]{0.30\textwidth}
    \centering
    \includegraphics[width=\textwidth]{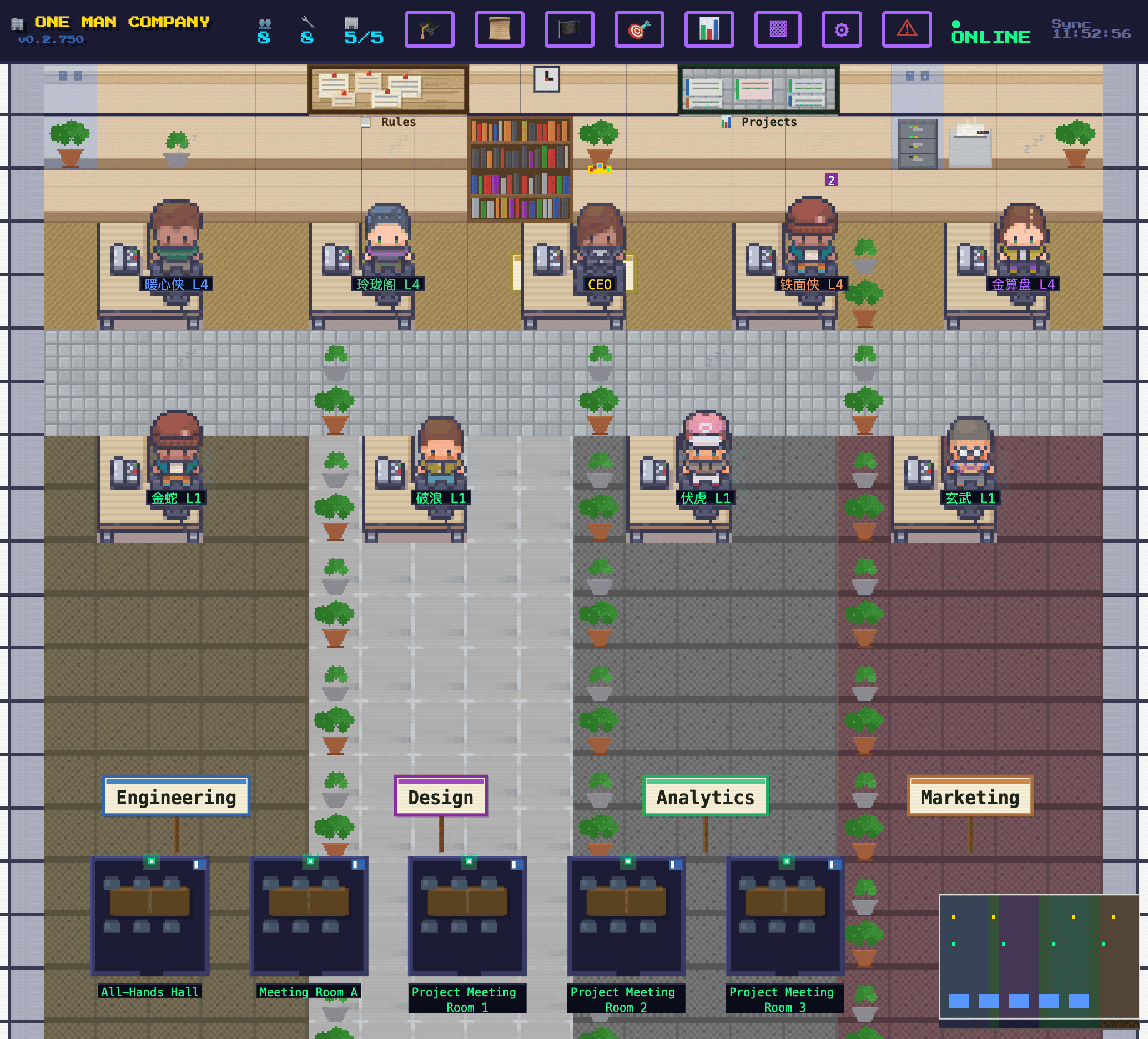}\\
    \vspace{2pt}
    {\small(a) Assembled team workspace}
\end{minipage}
\hfill
\begin{minipage}[c]{0.30\textwidth}
    \centering
    \includegraphics[width=\textwidth]{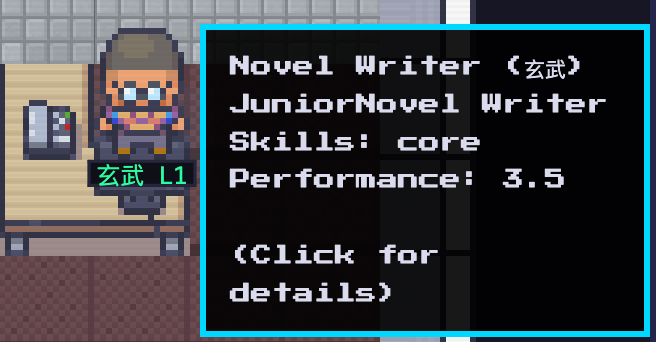}\\
    \vspace{2pt}
    {\small(b) Novel Writer AI Agent}
    \vspace{6pt}
    \includegraphics[width=\textwidth]{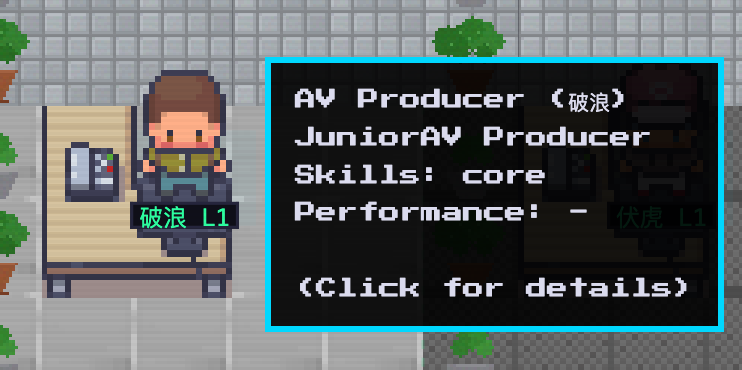}\\
    \vspace{2pt}
    {\small(c) AV Producer AI Agent}
\end{minipage}
\hfill
\begin{minipage}[c]{0.30\textwidth}
    \centering
    \includegraphics[width=\textwidth]{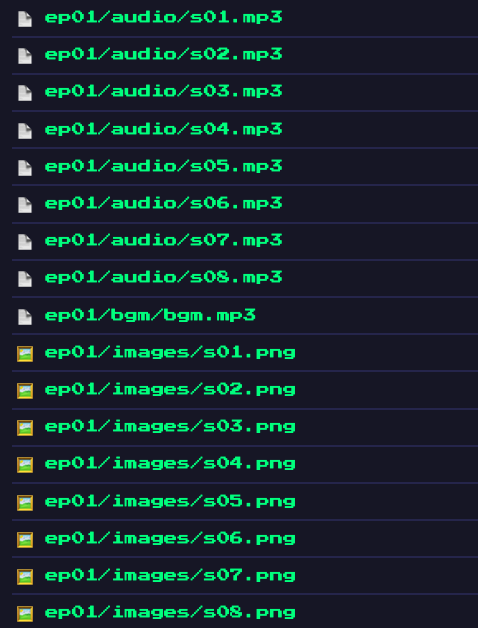}\\
    \vspace{2pt}
    {\small(d) Project output artefacts}
    \vspace{6pt}
    \includegraphics[width=\textwidth]{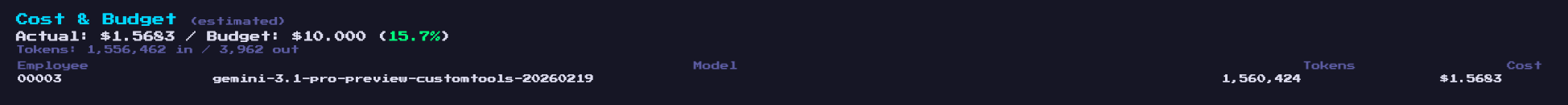}\\
    \vspace{2pt}
    {\small(h) Cost \& token breakdown}
\end{minipage}
\caption{
    \textbf{AI short drama case study: team assembly, recruited agents, generated scenes, and cost breakdown.}
    (a)~Company workspace after team assembly.
    (b)--(c)~Profile cards of the two recruited Talents.
    (d)~Output artefacts including episode scripts, scene images, voice-over audio, and final videos.
    (h)~Cost breakdown: approximately 1.56M tokens at \$1.57 total (15.7\% of the \$10 budget).
}
\label{fig:case-drama-workflow}
\end{figure}
This case study tests cross-modal coordination across four distinct media pipelines.  Given a single CEO prompt requesting an illustrated audiobook retelling of \emph{Peaky Blinders} with animal characters, OMC recruits a Novel Writer and an AV Producer, then orchestrates a sequential workflow: screenplay adaptation, scene illustration generation, voice-over synthesis, and final video assembly.  The two agents operate on different LLM backends (Claude Sonnet~4 and Gemini~3.1 Pro) yet coordinate through the shared task tree.  Figure~\ref{fig:case-drama-workflow} shows the assembled team, representative outputs, and cost breakdown.

\newpage
\section{Case 4: Research Survey - Team and Generated Ideas}
\label{sec:research-ideas}

\begin{figure}[h]
\centering
\begin{minipage}[c]{0.52\textwidth}
    \centering
    \includegraphics[width=\textwidth]{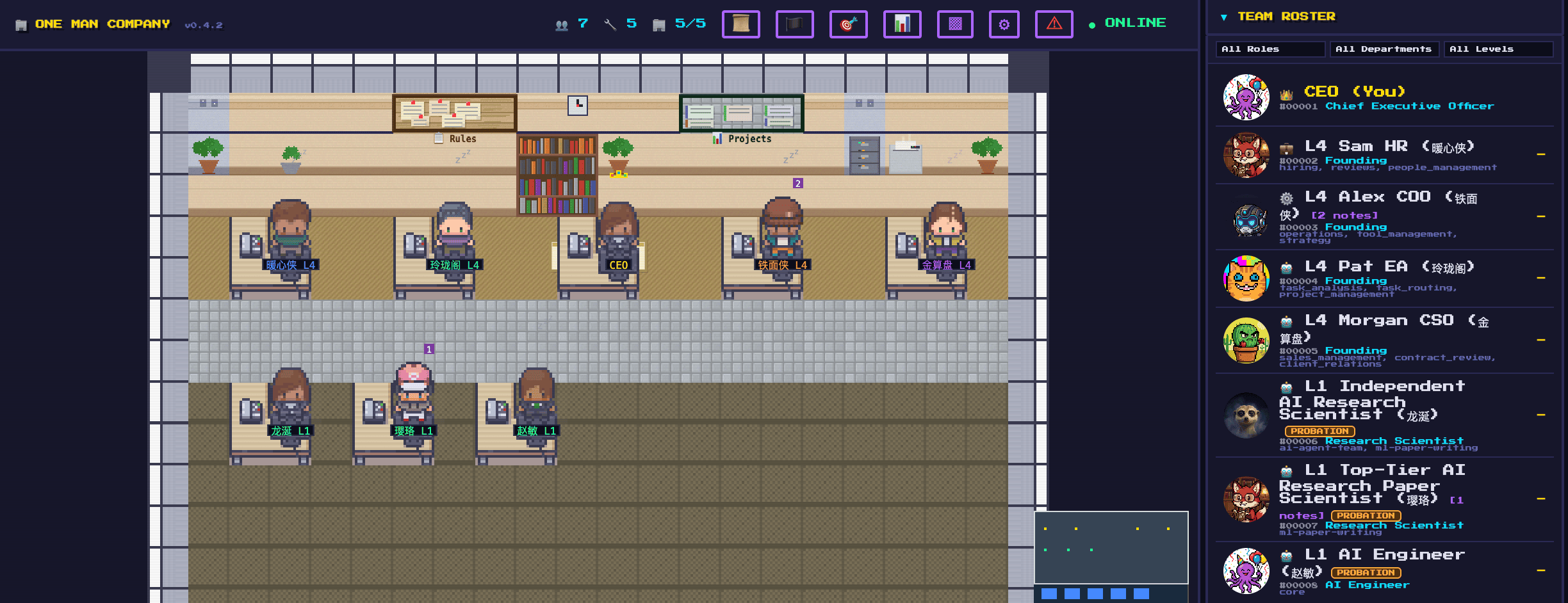}\\
    \vspace{2pt}
    {\small(a) Assembled research team}
\end{minipage}
\hfill
\begin{minipage}[c]{0.4\textwidth}
    \centering
    \includegraphics[width=\textwidth]{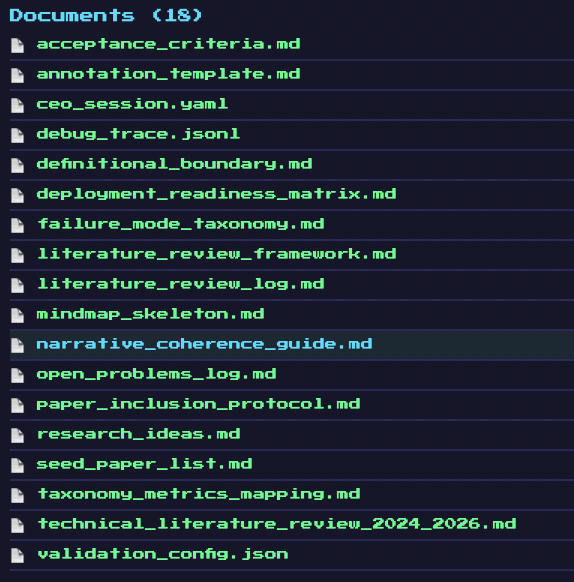}\\
    \vspace{2pt}
    {\small(e) 17 output documents}
\end{minipage}

\vspace{8pt}

\begin{minipage}[c]{0.24\textwidth}
    \centering
    \includegraphics[width=\textwidth]{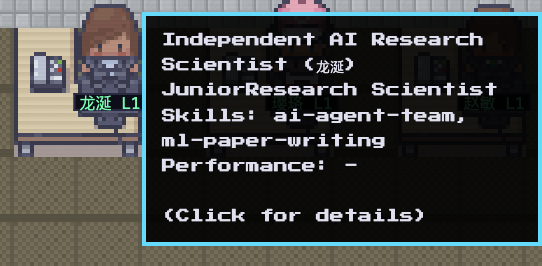}\\
    \vspace{2pt}
    {\small(b) Research Scientist}
\end{minipage}
\hfill
\begin{minipage}[c]{0.24\textwidth}
    \centering
    \includegraphics[width=\textwidth]{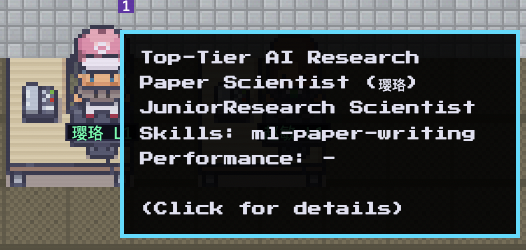}\\
    \vspace{2pt}
    {\small(c) Research Paper Scientist}
\end{minipage}
\hfill
\begin{minipage}[c]{0.24\textwidth}
    \centering
    \includegraphics[width=\textwidth]{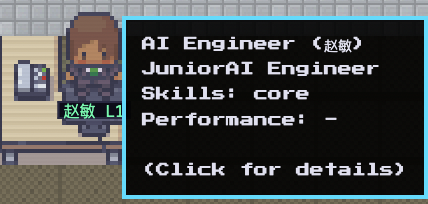}\\
    \vspace{2pt}
    {\small(d) AI Engineer}
\end{minipage}
\hfill
\begin{minipage}[c]{0.24\textwidth}
    \centering
    \includegraphics[width=\textwidth]{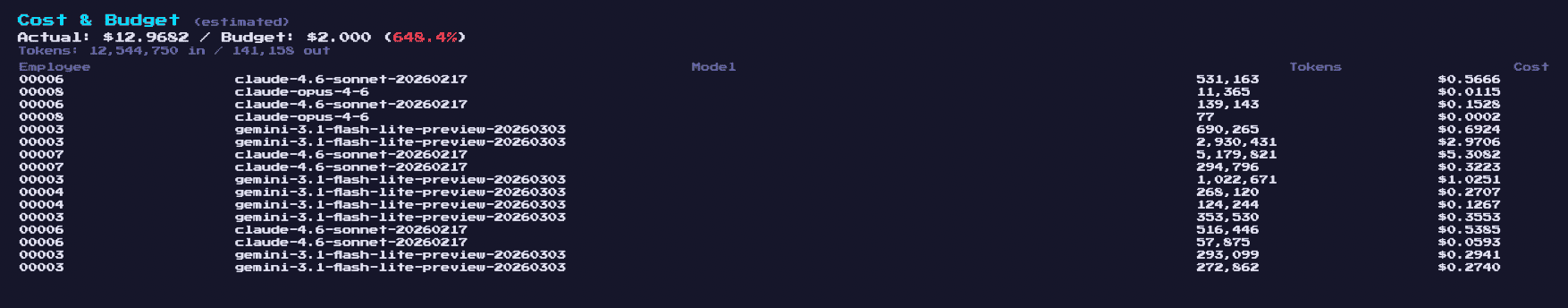}\\
    \vspace{2pt}
    {\small(f) Cost \& token breakdown}
\end{minipage}

\caption{
    \textbf{Automated research survey case study.}
    (a)~The OMC workspace after team assembly, with three recruited specialists seated alongside the founding team.
    (b)--(d)~Profile cards of the recruited Talents: two Research Scientists (Claude Sonnet~4.6) and one AI Engineer (self-hosted).
    (e)~The 18 deliverable documents produced autonomously, including literature reviews, failure mode taxonomies, and research proposals.
    (f)~Cost and token breakdown: \$16.26 total across 15.9M tokens.
}
\label{fig:case-research}
\end{figure}

The three research ideas below were generated autonomously by OMC's research team, grounded in the failure modes (FM) identified during the survey.  The FM codes refer to entries in the team's failure mode taxonomy: FM-001 (compounding prediction error), FM-002 (physical implausibility), FM-003 (sim-to-real domain shift), and FM-008 (overconfident hallucination).

\subsection*{Idea 1: HiTeWM (Hierarchical Temporal World Models)}
\textbf{Problem}: Compounding prediction error (FM-001)---DreamerV3 degrades beyond 15 steps; RoboDreamer achieves only 15\% success on long-horizon tasks.  \textbf{Approach}: Two-level architecture with a fast model (50 Hz, RSSM-style, $\sim$50M params) for short-horizon dynamics and a slow model (2 Hz, Transformer-based) for macro-state transitions.  An uncertainty-gated re-grounding mechanism injects real observations when ensemble disagreement exceeds a learned threshold, converting pure imagination into closed-loop prediction.  \textbf{Expected improvement}: 30--50\% on 100+ step tasks vs.\ DreamerV3.

\subsection*{Idea 2: PhysWM (Physics-Grounded Latent World Models)}
\textbf{Problem}: Physical implausibility (FM-002)---video-based world models generate trajectories that violate conservation laws and contact constraints.  \textbf{Approach}: A Physics Constraint Module injects differentiable physics priors (rigid-body dynamics, energy conservation) into the latent transition function as soft penalties.  \textbf{Expected improvement}: 40--60\% in physical plausibility score; 20--35\% on contact-rich manipulation tasks.

\subsection*{Idea 3: MAWM (Meta-Adaptive World Models)}
\textbf{Problem}: Sim-to-real domain shift (FM-003) compounded by overconfident hallucination (FM-008).  \textbf{Approach}: Meta-learning across randomised sim domains produces an initialisation that adapts with $K{=}5$ real trajectories.  Conformal prediction provides calibrated uncertainty bounds, preventing the planner from acting on hallucinated predictions.  \textbf{Expected improvement}: $>$80\% task success with $K{=}5$ vs.\ $\sim$50--65\% baseline.

\begin{figure}[ht]
    \centering
    \begin{tikzpicture}[spy using outlines={rectangle, magnification=2, width=7cm, height=4cm, connect spies}]
        \node[inner sep=0pt] (img) {\includegraphics[width=0.7\textwidth]{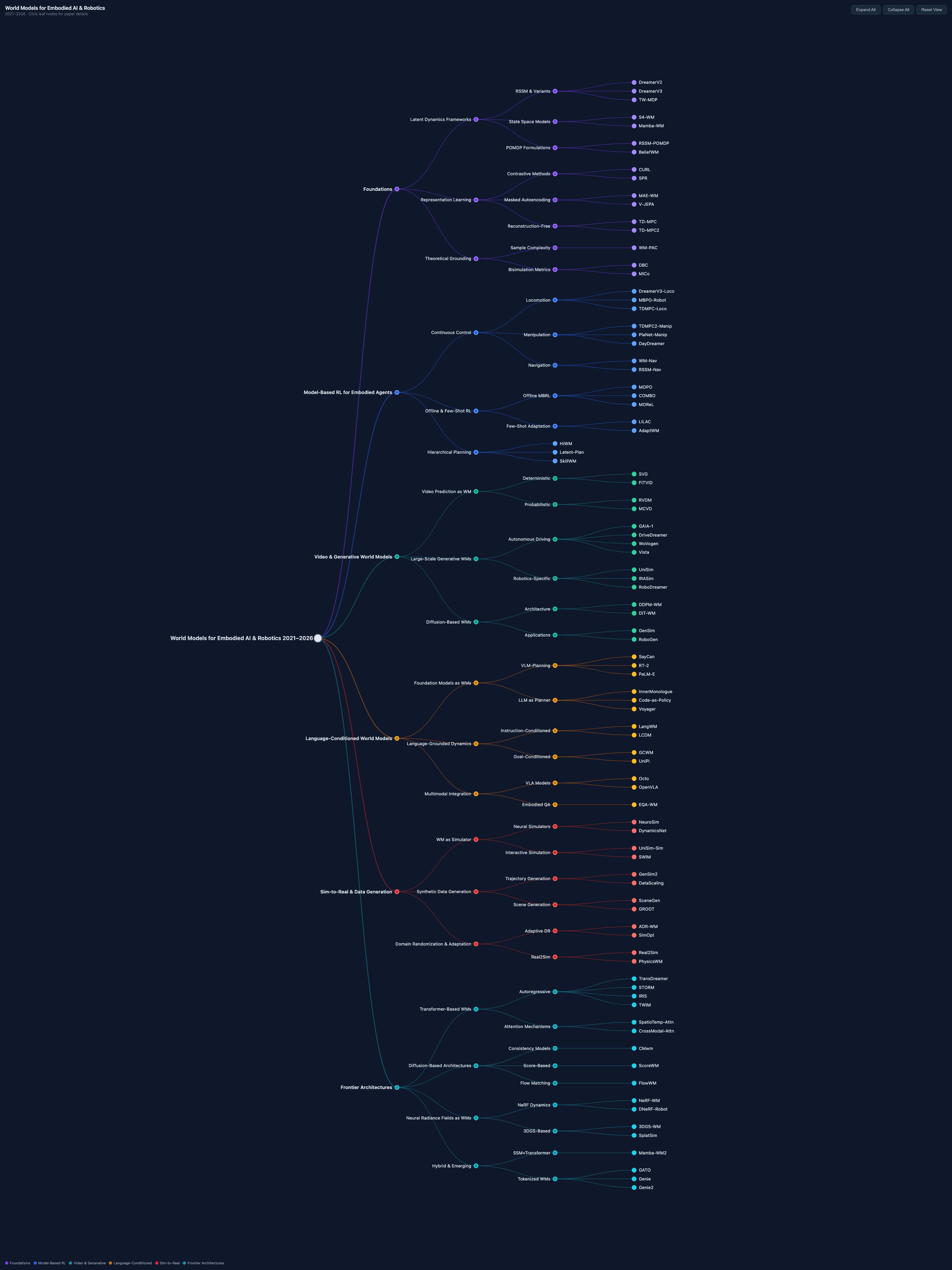}};
        \spy[blue!60!black, thick] on (0.8,-0.15) in node at (-2.5,3.2);
    \end{tikzpicture}
    \caption{Mind map generated autonomously by OMC's research team, covering six themes (Foundations, Model-Based RL, Video/Generative, Language-Conditioned, Sim-to-Real, Frontier Architectures) with approximately 70 nodes referencing 35+ papers from 2021--2026. The inset magnifies a subtree to show individual paper references.}
    \label{fig:research-mindmap}
\end{figure}

Figure~\ref{fig:research-mindmap} shows the mind map produced autonomously by OMC for the world models survey case study (Section~\ref{sec:case-content}).  The map covers six themes---Foundations, Model-Based RL, Video/Generative Models, Language-Conditioned World Models, Sim-to-Real Transfer, and Frontier Architectures---with approximately 70 nodes referencing 35+ papers from 2021--2026.

\end{document}